\documentclass[times, review, 10pt]{elsarticle}

\usepackage{graphicx}%
\usepackage{multirow}%
\usepackage{amsmath,amssymb,amsfonts}%
\usepackage{amsthm}%
\usepackage{mathrsfs}%
\usepackage[title]{appendix}%
\usepackage{xcolor}%
\usepackage{textcomp}%
\usepackage{manyfoot}%
\usepackage{booktabs}%
\usepackage{algorithm}%
\usepackage{algorithmicx}%
\usepackage{algpseudocode}%
\usepackage{listings}%
\usepackage{enumitem}

\usepackage{bbm}
\usepackage{soul, color, xcolor}
\usepackage{ulem}
\usepackage[left]{lineno}

\newcommand{\revise}[1]{{\color{black}#1}}

\usepackage[capitalize]{cleveref}
\crefname{section}{Sec.}{Secs.}
\crefname{appendix}{}{}
\Crefname{table}{Table}{Tables}
\Crefname{equation}{Eq.}{Eqs.}

\biboptions{sort&compress}

\begin{document}

\begin{frontmatter}

\title{Revisiting Out-of-Distribution Detection in Semantic Difference Space} 

\author{Xingming Long, Jie Zhang, Shiguang Shan, Xilin Chen} 

\affiliation{organization={State Key Laboratory of AI Safety, Institute of Computing Technology, Chinese Academy of Sciences, Beijing, China\\
University of Chinese Academy of Sciences},
state={Beijing},country={China}
}

\begin{abstract}
The primary goal of most existing out-of-distribution (OOD) detection works is to identify inputs with ``Semantic Shifts''.
However, the existing definition of ``Semantic Shift'' in previous studies remains vague, resulting in controversial choices of OOD samples in many benchmarks. Consequently, some benchmarks are too trivial for evaluating OOD detection, whereas others include OOD instances that are practically undetectable.
In this paper, we propose a formal definition of ``Semantic Difference Space'', which reflects the scope of a model's OOD detection capability after training convergence. The definition is grounded in an analysis of how the loss-based optimization shapes the model's perceptual toward OOD inputs, offering theoretical explanations for the controversies regarding the selection of OOD samples across existing benchmarks.
To validate the correctness of our analysis, we construct multiple datasets, primarily in the context of image classification, and we find that compared to different OOD detection methods, different OOD settings play a more decisive role in the detection results.
Finally, we propose a simple yet effective OOD Separability Probe (OSP) method to evaluate whether the existing OOD detection benchmarks are reasonable, identifying issues in some benchmarks and highlighting a few that are worth further exploration.
\end{abstract}



\end{frontmatter}

\section{Introduction}
\label{introduction}
Out-of-distribution (OOD) detection has gained increasing attention in recent years due to its crucial role in ensuring algorithmic reliability \cite{MSP_2017, OOD_D_Survey_2024}. While deep learning models excel on in-distribution (ID) samples, i.e., those belonging to the same classes as the training set, these models often struggle when encountering unseen OOD classes \cite{OOD_D_Survey_2024}. For example, a model trained on a fruit classification dataset may produce arbitrary and incorrect predictions when presented with an image of an animal. 
The primary objective of OOD detection is to enable models to identify inputs that fall outside the ID distribution and reject them, rather than blindly classifying them into existing ID classes, thus mitigating potential security risks.
Existing OOD detection research has made significant progress \cite{MSP_2017,RMDS_2021,NCI_2025}, achieving notable results on various OOD detection benchmarks \cite{OpenOOD_2022,ImageNetOOD_2024}.

\begin{figure}[h]%
\centering
\includegraphics[width=0.7\textwidth]{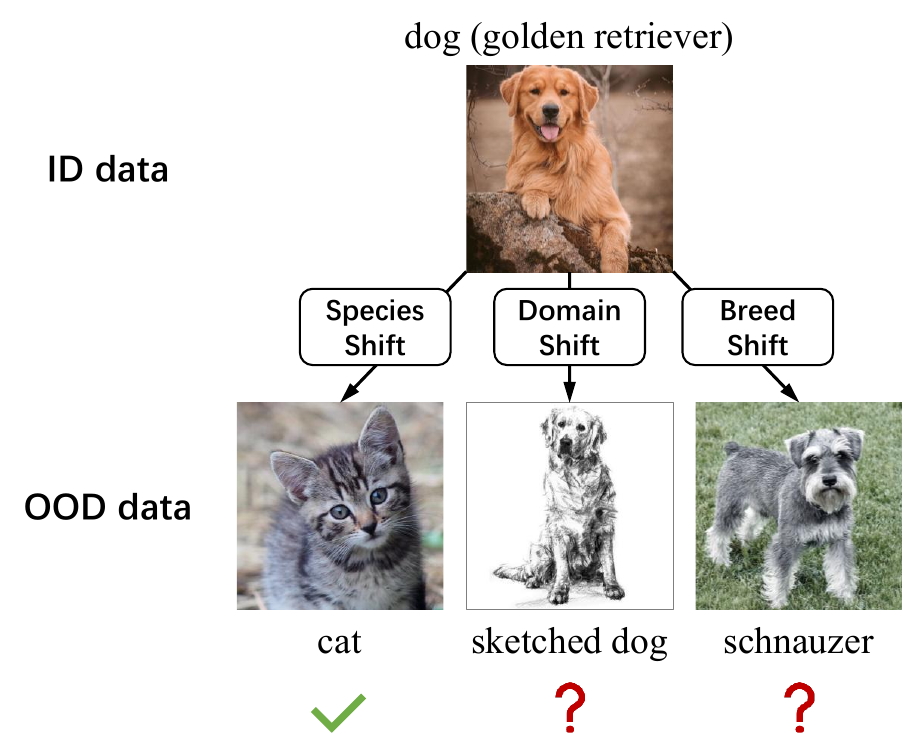}
\caption{An illustration of the ambiguity in the definition of ``Semantic Shift''. The shift in OOD data can occur across varying perspectives (such as the unseen sketch domain in the middle image) and granularities (such as the unseen breed in the right image), making it challenging to establish consistent criteria when constructing a benchmark.}
\label{fig:intro}
\end{figure}

However, we find that the current definition of OOD detection settings is ambiguous. In existing OOD literature, OOD samples are typically defined as data exhibiting Semantic Shifts relative to the training set \cite{OOD_D_Survey_2024}, yet the scope of ``Semantic Shift'' remains poorly defined.
While ``Semantic Shift'' can arise across different perspectives and granularities, there is no clear consensus on which of these shifts warrant treating a sample as OOD.
For example, as shown in \cref{fig:intro}, given an ID dataset that includes the class ``dog'', should a sketch of dog or a previously unseen breed be considered as OOD data?
The ambiguity is reflected in the inconsistent criteria for partitioning OOD data across different benchmarks, which introduce varying subjective standards for categorizing OOD samples \cite{OpenOOD_2022,ImageNetOOD_2024}. Consequently, some benchmarks are too trivial for evaluating OOD detection, whereas others include OOD instances that are practically undetectable. Moreover, we also find that many benchmarks include OOD settings referred to as ``Far-OOD''. For example, using a texture dataset as OOD data for an object classification ID dataset. Whether it is meaningful to use a completely different type of data as the OOD data remains a subject of debate. This raises two fundamental questions: what determines the theoretical scope of a model's OOD detection capability, and how can we systematically assess whether a given benchmark aligns with this scope?

In this paper, we first formally define the ``Semantic Difference Space'' to provide a clear sight on the scope of the model's OOD detection capability, which is grounded in the theoretical analysis under both linear and nonlinear assumptions. The analysis shows how the model's perceptual toward OOD inputs is shaped during the training convergence process.
To further validate our findings with practical experiments, we construct multiple OOD detection datasets, primarily in the context of image classification. The experiments on each dataset show that the detection results are significantly affected by using OOD settings constructed through different perspectives and granularities.
Finally, we introduce the OOD Separability Probe (OSP) method to measure the separability between ID and OOD samples in existing benchmarks and to assess the reasonableness of the settings in these benchmarks. The Semantic Difference Space framework provides a theoretical answer to what subspace a model learns to be sensitive to, while OSP operationalizes this insight as a practical benchmark auditing tool. Based on the results, we identify benchmarks that are unsuitable for OOD detection, as well as promising ones that worth further research.

Our main contributions can be summarized as follows:
\begin{itemize}
    \item We provide a formal definition of ``Semantic Difference Space'' based on theoretical analysis, which reflects the scope of a model's OOD detection capability and offers a reference criterion for the selection of OOD samples in a benchmark.
    
    \item To validate our analysis, we construct multiple OOD detection datasets that demonstrate how the detection performances are determined by the OOD settings constructed with different perspectives and granularities.
    
    \item We propose a simple yet effective OOD Separability Probe (OSP) method to evaluate whether existing benchmarks are reasonable for OOD detection. The results reveal flaws in some benchmarks and identify others that warrant further study.
\end{itemize}

\section{Related Work}
\label{related work}
\subsection{OOD Detection Method}
Existing OOD detection methods can be grouped into three main categories \cite{OpenOOD_2022}: post-hoc methods, training-time regularization and training with outlier exposure.

The majority of works focus on post-hoc methods \cite{MSP_2017,ODIN_2018,EBO_2020}, which fixed the original model parameters and utilize the internal variables or outputs of the model to assess whether a sample is OOD. 
Many researchers utilize the model's output logits \cite{MSP_2017,SCALE_2024}, assuming that the logits for OOD data exhibit a more uniform distribution across different classes compared to ID data. Another line of typical approaches uses the intermediate features to estimate the probability that the test data belongs to ID distribution, either by modeling the ID data distribution with a probabilistic model \cite{AD_GMM_2018} or by calculating the distance between test samples and ID data \cite{KNN_2022,STOOD-X_2026}. Additionally, some methods identify OOD samples by detecting anomalies in the gradient or internal activations of the model \cite{GradNorm_2021,ReAct_2021}. Retrieval-augmented prompts are also utilized for OOD detection \cite{RAP_2026}. Recent studies further explore OOD detection with vision-language models, for example by dynamically evolving cross-modal proxies \cite{CoEvo_2026} or by using optimal transport to measure both semantic and distributional discrepancies between images and ID labels \cite{OT-DETECTOR_2025}.

The second major category of methods involves training-time regularization \cite{RotPred_2019,PALM_2024}, which alter the training process to improve the model's ability to predict OOD classes. 
Various regularization losses are added during training to enhance the model's ability to distinguish OOD samples, such as contrastive loss \cite{CSI_2020} and logit normalization \cite{LogitNorm_2022}. Recent advancements in contrastive learning further refine feature representations by introducing multiple prototypes to capture fine-grained semantics \cite{prototype_2025} or employing a class-aware mapping to strengthen feature discriminability \cite{CAFCL_2025}. Other works refer to different strategies to generate OOD-like samples, using these samples to help the model more accurately delineate the boundary between ID and OOD data \cite{VOS_2022,NPOS_2023}. One interesting work leverage the super category of the training data, performing OOD detection within each super category to enhance the model's detection capability \cite{MOS_iNaturalist_2021}.

The third approach category incorporates external OOD data during training \cite{OE_2019,SCOOD_UDG_2021}, helping the model distinguish between ID and OOD data throughout the training phase. Since this paper focuses primarily on studying the impact of training data on the model's OOD detection capability, we exclude these methods as they involve extra information during training.

\subsection{OOD Detection Benchmark}
Early OOD detection benchmarks treat differences in data labels as Semantic Shifts \cite{MSP_2017,OpenOOD_2022}, using one dataset as the ID data for training and selecting another dataset that does not contain any ID classes as the OOD data for testing. For example, a common setting is to use ImageNet-1K \cite{ImageNet_2009} as ID data to train the model, and then use datasets such as ImageNet-O \cite{ImageNet-O_2021} and OpenImage-O \cite{ViM_OpenImage-O_2022}, which have different labels from ImageNet-1K, as OOD data (denoted as Near-OOD), or use datasets like Texture \cite{Texture_2011}, MNIST \cite{MNIST_2012}, and SVHN \cite{SVHN_2011}, which are completely different types from ImageNet-1K, as OOD data (denoted as Far-OOD).

Later researches reveal that data with different labels can also contain similar semantic contents \cite{NINCO_2023,ImageNetOOD_2024}, leading to the adoption of heuristic or rule-based methods to filter out marginal ID samples from the OOD dataset. Going a step further to reduce label dependence, \cite{GBench_2025} proposes a Gscore indicator to enable unsupervised OOD detection evaluation. More recent studies suggest that semantic shift is continuous \cite{COOD_2023,ISOOD_2024}, splitting the test set into multiple subsets with varying degrees of shifts to analyze the model's performance variations. Another work identifies the impact of covariate shift on OOD detection and incorporates covariate shifted datasets into their benchmarks \cite{FSOOD_2023}. Compared with the previous works, the constructed datasets in this paper no longer contain a single training task, enabling a more comprehensive investigation into how different perspectives and granularities in OOD settings impact OOD detection performance.

\subsection{OOD Detection Theory}
Compared to the research on OOD detection methods and benchmarks, fewer works study the theory of OOD detection.
Several works on OOD detection theory explore factors that can enhance OOD detection performance. The labels of ID data can significantly improve OOD detection performance, as shown in \cite{Theory_ID_label_2024}, while the additional unlabeled data is also proved to be beneficial for the OOD detection task \cite{Theory_OE_unlabel_2024}.
Another line of research aims to provide guarantees or bounds for the OOD detection task. One study provides guarantees for a specific detection score (Mahalanobis distance) under Gaussian mixture assumptions \cite{Theory_MixGaussian_2022}. The theory in \cite{Teory_leanable_2022} focuses on the PAC learnability of OOD detection in different scenarios.
Compared to these theoretical works that focus on guarantees for specific detection scores \cite{Theory_MixGaussian_2022} or PAC learnability conditions for OOD detection algorithms \cite{Teory_leanable_2022}, we address a different and complementary question: what subspace of the input space does a convergent classifier become sensitive to, and how does this inherent detection scope, determined by the ID categorization, constrain which OOD shifts are detectable in principle?


\section{Theoretical Analysis}
\subsection{Preliminary}
\label{Preliminary}
Below, we provide the notions for Out-of-Distribution (OOD) detection as described in most existing works \cite{Theory_MixGaussian_2022,Theory_ID_label_2024}.

\textbf{Labeled ID distribution.} \textit{Let the input space $\mathcal{X}$ be a d-dimensional space $\mathbb{R}^d$. The label space for ID data is denoted as $\mathcal{Y}_I=\{y_1,...,y_k\}$. The ID data is independently and identically sampled from the joint distribution $(\boldsymbol{x},y) \sim \mathcal{P}_I$ defined over $\mathcal{X} \times \mathcal{Y}_I$.}

\textbf{OOD distribution.} \textit{The input of the OOD distribution is also from the space $\mathcal{X}$ but with OOD label $\mathcal{Y}_O=\{y_o\}$, where $y_o \notin \mathcal{Y}_I$. The OOD data is drawn from the joint distribution $(\boldsymbol{x},y) \sim \mathcal{P}_O$ defined over  $\mathcal{X} \times \mathcal{Y}_O$. For OOD detection, only the input $\boldsymbol{x}$ is available for use by the models.}

However, the definition above lacks connections between the input and the label. In real-world tasks, data sharing the same label inevitably exhibits common features, such as elephants having long trunks and cars having round wheels. Therefore, we will further introduce constraints between inputs and labels in the following analysis, first in the linear case and then in more general cases.

\subsection{Linear Space Analysis}
\label{sec:linear_analysis}
We first conduct the analysis in the relatively simple linear condition. As we assume that data from the same class should share similar inputs, we explicitly model the input of each ID class as a Gaussian distribution following the previous setup in \cite{Theory_MixGaussian_2022}. The definition is as follows:

\textbf{Data Distribution in Linear Space.}
\label{section:OOD_detection_in_Gaussian}
\textit{For the class-conditional ID distribution $(\boldsymbol{x},y_i) \sim \mathcal{P}_I|y_i, i\in\{1,...,k\}$, we assume the input follows a Gaussian distribution $\boldsymbol{x} \sim \mathcal{N}(\boldsymbol{\mu}_i,\boldsymbol{I})$, where $\boldsymbol{\mu}_i$ denotes the class center. The centers of each class are linearly separable.}

We next provide our formal definition of ``Semantic Difference Space'' in such case. The definition is as follows:

\textbf{Definition 1.} \textit{The Semantic Difference Space $\mathcal{S}$ based on the ID distribution is defined as}
\begin{equation} \begin{aligned}
\mathcal{S} = span(\{\boldsymbol{\mu}_1-\boldsymbol{\mu}_2,...,\boldsymbol{\mu}_{k-1}-\boldsymbol{\mu}_{k}\}).
\label{eq:sem_space}
\end{aligned} \end{equation}

It can be seen that the Semantic Difference Space depends on the categorization of the classes in the ID data, as it is determined by the center $\boldsymbol{\mu}_i$ of each class. The Semantic Difference Space is constructed based on inter-class differences, ensuring that it only captures the distinct characteristics of each class.

We then define the corresponding complementary space as follows to ensure the completeness of the definition:

\textbf{Definition 2.} \textit{The Covariate Space $\mathcal{C}$ is defined using direct sum decomposition}
\begin{equation} \begin{aligned}
\mathcal{X} &= \mathcal{S} \oplus \mathcal{C}, \\
\text{i.e., } \mathcal{C} = \{ c \in \mathcal{X} &\mid \forall s \in \mathcal{S}, \langle s, c \rangle = 0 \},
\label{eq:cov_space}
\end{aligned} \end{equation}
\textit{where $\langle \cdot, \cdot \rangle$ denotes the inner product.}

It is worth emphasizing that the Covariate Space is determined by the Semantic Difference Space, meaning the Covariate Space also indirectly depends on the categorization within the ID distribution. For instance, in object detection, the image background is considered part of the Covariate Space, whereas it is not in scene recognition tasks.
Note that the rank of $\mathcal{S}$ is at most $\min(k-1, d)$: when class centers are nearly collinear, $\mathcal{S}$ becomes low-dimensional and the detectable OOD scope is correspondingly restricted; conversely, when $k \gg d$, $\mathcal{S}$ may span the entire $\mathcal{X}$, leaving no Covariate Space and thus no dimension in which the model is invariant.

After providing the definition of Semantic Difference Space, we conduct theoretical analysis to explore how the model's OOD detection capability is constrained in this subspace during training convergence. 
We employ a linear classifier $f$ with a weight matrix $\boldsymbol{W}$ as the model to study its perception capability across different subspaces.

Under the linear condition, we derive the following important proposition. The proof can be found in Appendix A.

\textbf{Proposition 1.} \textit{Under mild training assumptions, the weight matrix $\boldsymbol{W}$ of the linear classifier $f$, after being trained on the ID distribution, can always be decomposed into a simplified weight matrix $\tilde{\boldsymbol{W}}$ and an orthogonal matrix $\boldsymbol{Q}$. 
The last $d - r$ columns of $\tilde{\boldsymbol{W}}$ are all zero vectors, and the first $r$ rows of $\boldsymbol{Q}$ form an orthogonal basis for the Semantic Difference Space $\mathcal{S}$. $r$ is the rank of the Semantic Difference Space $\mathcal{S}$.}
\begin{equation} \begin{aligned}
\boldsymbol{W} &= \tilde{\boldsymbol{W}} \boldsymbol{Q}, \\
s.t. \quad \tilde{\boldsymbol{W}}_{:,r+1} &= ... = \tilde{\boldsymbol{W}}_{:,d} = \boldsymbol{0}, \\
span(\{\boldsymbol{Q}_{1,:}&,...,\boldsymbol{Q}_{r,:}\}) = \mathcal{S}.
\label{eq:pro_1}
\end{aligned} \end{equation}

The proposition shows that after orthogonal transformation, only the weight matrix corresponding to the Semantic Difference Space is non-zero, while the rest part of the matrix corresponding to the Covariate Space will converge to zero. This implies that the changes of the input vector within the Covariate Space no longer affects the classifier's output once training has converged.

Using the above proposition, the following theorem can be easily derived, with its proof provided in Appendix A.

\textbf{Theorem 1.} \textit{For any two classes of data with Gaussian distributions $\mathcal{N}(\boldsymbol{\mu}_a,\boldsymbol{I})$ and $\mathcal{N}(\boldsymbol{\mu}_b,\boldsymbol{I})$ within the input space $\mathcal{X}$, if the difference between their class centers $\boldsymbol{\mu}_a$ and $\boldsymbol{\mu}_b$ is $\mathbf{0}$ in the Semantic Difference Space $\mathcal{S}$, they are indistinguishable to the linear classifier $f$ trained on the ID distribution, regardless of how different they are in the Covariate Space $\mathcal{C}$}
\begin{equation} \begin{aligned}
&\text{if} \quad proj_\mathcal{S}(\boldsymbol{\mu}_a-\boldsymbol{\mu}_b) = \mathbf{0}, \\
KL(&f(\mathcal{N}(\boldsymbol{\mu}_a,\boldsymbol{I})) || f(\mathcal{N}(\boldsymbol{\mu}_b,\boldsymbol{I}))) = 0.
\label{eq:theorem_1}
\end{aligned} \end{equation}

The theorem provides a necessary condition for the model to distinguish between ID and OOD data: the ``Semantic Shift'' of the OOD data must exhibit within the defined ``Semantic Difference Space''.

\subsection{More General Analysis}
\label{sec:non-linear_analysis}
After analyzing the case in the linear space, we also aim to analyze how the training impacts the model's perception of inputs under more general non-linear conditions. 

Following the idea in Definition 1, our goal is to define the Semantic Difference Space in a way that can capture the differences between the input of different ID classes.

Based on this idea, we implicitly define the Semantic Difference Space $\mathcal{S}$ using the objective function $g(\cdot)$, which truly reflects the relationship between the input space $\mathcal{X}$ and the output ground truth $\mathcal{Y}_I$ within the ID distribution $\mathcal{P}_I$. The formal definition is as follows:

\textbf{Definition 3.} \textit{\revise{Given a continuously differentiable objective function $g(\cdot)$ of the ID distribution}, the Semantic Difference Space $x_s=[s_1,s_2,...,s_r]^\top \in \mathbb{R}^{r}$ is defined as the input space of the function $g(\cdot)$, where $g: x_s \mapsto y$, satisfying that the partial derivative of the ground truth output with respect to any input variable is not identically zero $\frac{\partial g}{\partial s_i} \not\equiv 0$. The Covariate Space $x_c=[c_1,c_2,...,c_{d-r}]^\top \in \mathbb{R}^{d-r}$ is then constructed by the rest inputs of the model function $f(\cdot)$, where $f: (x_s, x_c) \mapsto \hat{y}$.}

We make requirements on the partial derivative $\frac{\partial g}{\partial s_i} = \underset{\Delta s_i \to 0}{\lim} \frac{g(x_s+\Delta s_i \mathbf{e}_i)-g(x_s)}{\Delta s_i} \not\equiv 0$, indicating the differences of input within the Semantic Difference Space lead to the changes in output value. In this case, the Semantic Difference Space is implicitly constructed based on the differential relations among data, whereas in the linear case it is based on inter-class differences. \revise{Here $g(\cdot)$ can be understood as a latent decision function in the discrete classification task. It outputs a $k$-dimensional vector of class scores and the discrete class label is obtained as $y_{\mathrm{dis}}=\underset{{i\in\{1,\ldots,k\}}}{\arg\max} g(x_s)$.} It describes how perceptual features map to category membership. For example, $g(\cdot)$ is sensitive to changes in breed-specific features, such as ear shape and coat texture, but invariant to background or lighting, so $\mathcal{S}$ captures the former while $\mathcal{C}$ captures the latter.

After providing the definition of Semantic Difference Space under the more general non-linear condition, we can analyze the training process based on the definition. During the training of the model function $f(\cdot)$, the objective is to minimize the discrepancy between the model's predicted output $\hat{y}$ and the ground truth $y$ over the input distribution $\mathcal{P}_I$.

We then derive the following theorem. The proof is provided in Appendix A.

\textbf{Theorem 2.} \textit{If the inputs are independent on the ID distribution $\mathcal{X}$ and the model converges on that distribution, i.e., for any loss upper bound $\epsilon_{loss}>0$, if the model satisfies that $E_{\mathcal{X}}L(f,g) < \epsilon_{loss}$, then within an input region $\Delta \mathcal{X}=\{(x_s,x_c)| a_{s,i} \leq s_i \leq b_{s,i}, a_{c,j} \leq c_j \leq b_{c,j}\}$ with probability $P_{\Delta \mathcal{X}}=\int_{\Delta \mathcal{X}}p(x)dx$, for any input $c_j$ within the Covariate Space satisfies that its probability density is greater than a positive constant $p(c_j) \geq p_m$ and \revise{the partial derivative $\frac{\partial f}{\partial c_j}$ is constant throughout the analyzed region}, the partial derivative of the model function $f(\cdot)$ with respect to that input $c_j$ will have an upper bound:}
\begin{equation} \begin{aligned}
|\frac{\partial f}{\partial c_j}| \leq \frac{4\epsilon_{loss}}{p_m P_{\Delta \mathcal{X}} k_{\Delta \mathcal{X}} (b_{c,j}-a_{c,j})^2},
\end{aligned} \end{equation}
\textit{where $L(\cdot,\cdot)$ is the loss function, $p(\cdot)$ denotes the probability density function and $k_{\Delta \mathcal{X}}$ is a constant coefficient that ensures the loss function is meaningful on the region $\Delta \mathcal{X}$ (formally defined in Appendix A Definition 4).}

This theorem proves that as training progresses and the loss $L(f,g)$ decreases, the upper bound of the partial derivatives $\frac{\partial f}{\partial c_j}$ will also decrease. Ultimately, as the training converges and the loss approaches zero, the established bound guarantees that the partial derivative converges to zero:
\begin{equation}
\lim_{E_{\mathcal{X}}L(f,g) \to 0} \frac{\partial f}{\partial c_j} = 0.
\end{equation}
This strictly aligns with the result obtained in Proposition 1 under linear conditions. At this point, the shifts within the Covariate Space no longer influence the model's output, and we can again obtain the conclusion in Theorem 1.

\subsection{Discussion}
All the preceding analyses indicate that a model's OOD detection capability is intrinsically constrained by the Semantic Difference Space learned from the ID distribution. In practice, the ID categorization is typically fixed by the primary task. Therefore, to construct a meaningful OOD detection benchmark, one should explicitly identify the Semantic Difference Space determined by the ID data and select OOD samples whose distributional shift from the ID data lies primarily within this subspace. This ensures the evaluation tests the model's genuine sensitivity to OOD anomalies, rather than being noised by covariate factors that the model has learned to ignore.

For instance, when a model is trained on ImageNet-1K \cite{ImageNet_2009}, the inclusion of numerous fine-grained dog breed categories ensures that breed-related features are captured within Semantic Difference Space. Consequently, introducing an unseen breed as OOD data is acceptable. In contrast, for a model trained on CIFAR-10 \cite{CIFAR_2009}, which lacks such fine-grained labels, breed-specific features reside outside Semantic Difference Space. Treating an unseen breed as OOD in this scenario would be inappropriate, as the model inherently lacks the representational capacity to meaningfully detect this specific shift.

\subsection{Theoretical Scope and Limitations}
The linear analysis in Theorem 1 relies on Gaussian class-conditional distributions and linear separability as simplifying assumptions to enable tractable derivations. Theorem 2 removes the Gaussian and linear-classifier assumptions, but it still depends on additional regularity conditions, such as input independence and \revise{constant partial derivative within the analyzed region}. Therefore, Theorem 2 should be viewed as a conceptual extension from the linear case rather than a complete characterization of modern deep network training, which may involve non-convex optimization, batch normalization, and implicit regularization. Despite these limitations, our empirical results across ResNet-50 and ViT-based architectures are consistent with the conceptual predictions of the framework.

\section{OOD Separability Probe}
\label{sec:OSP}
\begin{figure}[t]%
\centering
\includegraphics[width=0.7\textwidth]{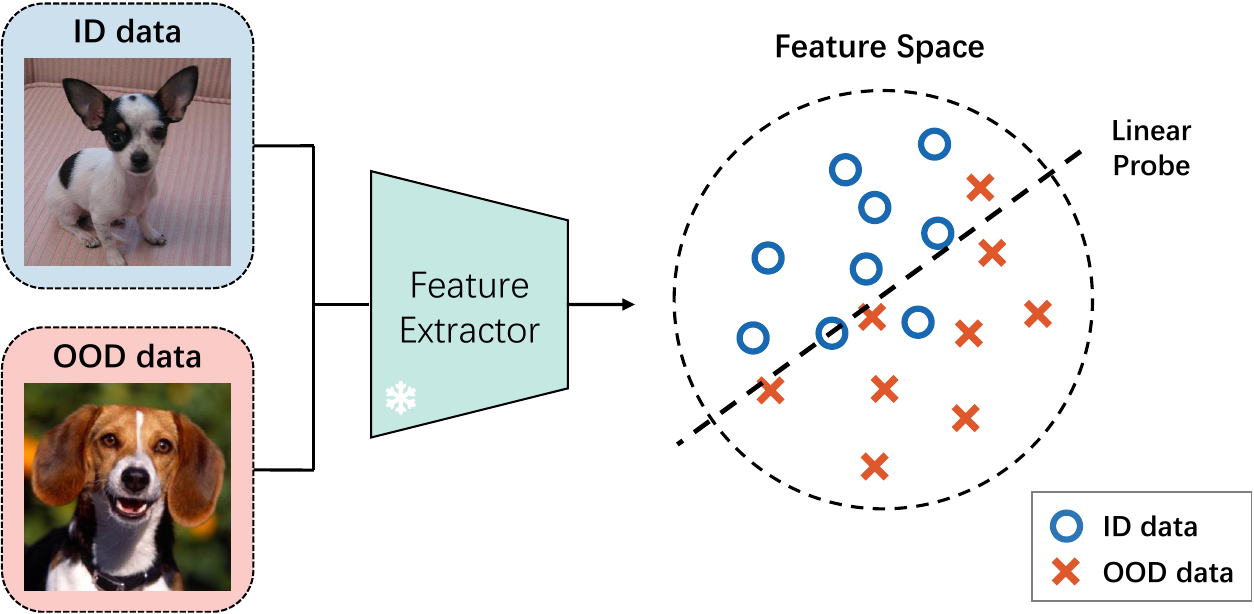}
\caption{An illustration of the OOD Separability Probe (OSP) method, which measures the OOD Separability between ID and OOD data within the model's feature space.}
\label{fig:probe}
\end{figure}

For more practical purposes, we introduce the OOD Separability Probe (OSP) method to assess the settings in existing benchmarks, evaluating their suitability for OOD detection tasks.
Specifically, we introduce a linear probe into the feature space of the converged model to measure the OOD Separability between OOD and ID data under a specific OOD settings, as shown in \cref{fig:probe}.

With the parameters of the model's feature extraction part fixed, the probe is optimized on a small subset of ID and OOD data to distinguish between their features. The probe's output represents the probability that the input is an OOD sample, and the optimization objective is to ensure that OOD samples produce higher outputs while ID samples produce lower outputs, as formally defined below:
\begin{equation} \begin{aligned}
\max_{f_{prob}} \left\{ \right. &\mathbb{E}_{\boldsymbol{x} \sim \mathcal{P}_I}  \log\left(1-f_{prob}(\phi(\boldsymbol{x}))\right) \\
+ &\mathbb{E}_{\boldsymbol{x} \sim \mathcal{P}_O} \log\left(f_{prob}(\phi(\boldsymbol{x}))\right) \left. \right\},
\label{eq:prob}
\end{aligned} \end{equation}
where $\phi$ denotes the feature extractor of the model and $f_{prob}$ refers to the linear probe.

We measure the OOD Separability using the classification AUROC of the optimized linear probe on OOD and ID data, which reflects the probability that the probe output of a random OOD sample is higher than that of a random ID sample, as follows:
\begin{equation} \begin{aligned}
&S(\mathcal{P}_I, \mathcal{P}_O) = \\
&\mathbb{E}_{\boldsymbol{x}_I \sim \mathcal{P}_I, \boldsymbol{x}_O \sim \mathcal{P}_O}P\left(f_{prob}(\phi(\boldsymbol{x}_O))>f_{prob}(\phi(\boldsymbol{x}_I))\right),
\label{eq:auroc}
\end{aligned} \end{equation}
where $S(\mathcal{P}_I, \mathcal{P}_O)$ represents the OOD Separability between the ID distribution $\mathcal{P}_I$ and the OOD distribution $\mathcal{P}_O$ within the feature space of the model trained on $\mathcal{P}_I$.
We choose a linear probe because mainstream architectures (e.g., ResNet, ViT) use a linear classification head, so a linear probe directly reflects the separability that the model itself can exploit.

Once the OOD Separability is measured under a specific OOD setting in a benchmark, \textbf{it can serve as a direct indicator of the OOD setting's difficulty}, thereby allowing us to assess the rationality of the chosen OOD setting. If the OOD Separability is too high, it indicates that the benchmark setting is too simple for the OOD detection task, and research on it may not significantly contribute to improving the effectiveness of OOD detection methods. On the other hand, if the OOD Separability is too low, it suggests potential issues with the choice of OOD samples. These OOD samples may not exhibit shifts within the Semantic Difference Space, which makes them difficult for the model to detect in principle; results under such settings should therefore be interpreted with caution.

\section{Datasets}
\label{dataset}
To investigate the impact of OOD settings on a model's OOD detection performance, we construct one synthetic dataset and two specially designed real-world image datasets. In this section, we provide a detailed introduction to the construction and testing protocols of these datasets.

\subsection{Synthetic Dataset}
We first design a simple experiment based on a synthetic dataset to validate the correctness of our theoretical analysis.
The dataset adopts a four-dimensional vector space as the input space $\mathcal{X}$. The first two dimensions are referred to as the Semantic Difference Space, while the remaining two are designated as the Covariate Space during data generation.

We set four ID classes for training and identify the class centers as follows:
\begin{equation} \begin{aligned}
\boldsymbol{\mu}_1 &= [\sigma,\sigma,\sigma,\sigma]^\top,\quad
\boldsymbol{\mu}_2 = [\sigma,-\sigma,\sigma,\sigma]^\top,\\
\boldsymbol{\mu}_3 &= [-\sigma,\sigma,\sigma,\sigma]^\top,\quad
\boldsymbol{\mu}_4 = [-\sigma,-\sigma,\sigma,\sigma]^\top,
\label{eq:exp_vectors}
\end{aligned} \end{equation}
where $\sigma$ is a constant that determines the position of the class centers.

We set different OOD classes for testing. Specifically, the OOD class centers $\boldsymbol{\mu}_o$ are constructed as follows:
\begin{equation} \begin{aligned}
&\boldsymbol{\mu}_o = \boldsymbol{s}_o + \boldsymbol{c}_o,\\
&\boldsymbol{s}_o \in \{[\sigma,\sigma,0,0]^\top, [0,\sigma,0,0]^\top\},\\
&\boldsymbol{c}_o \in \{[0,0,\sigma,\sigma]^\top, [0,0,-\sigma,\sigma]^\top, [0,0,-\sigma,-\sigma]^\top\}.
\label{eq:exp_shift_vectors}
\end{aligned} \end{equation}

The training data is randomly sampled from the ID distribution $\mathcal{N}(\boldsymbol{\mu}_i,\boldsymbol{I})$ with label $i\in\{1,2,3,4\}$. The test data is sampled from one of the OOD distribution $\mathcal{N}(\boldsymbol{\mu}_o,\boldsymbol{I})$ as well as from one of the ID distribution $\mathcal{N}(\boldsymbol{\mu}_1,\boldsymbol{I})$.

We randomly generate a fixed orthogonal matrix $\boldsymbol{Q}$ and apply it to each sampled data before put them into the model. This scrambling makes the Semantic Difference Space and the Covariate Space not explicitly distinguishable.

\begin{figure}[t]%
\centering
\includegraphics[width=0.7\textwidth]{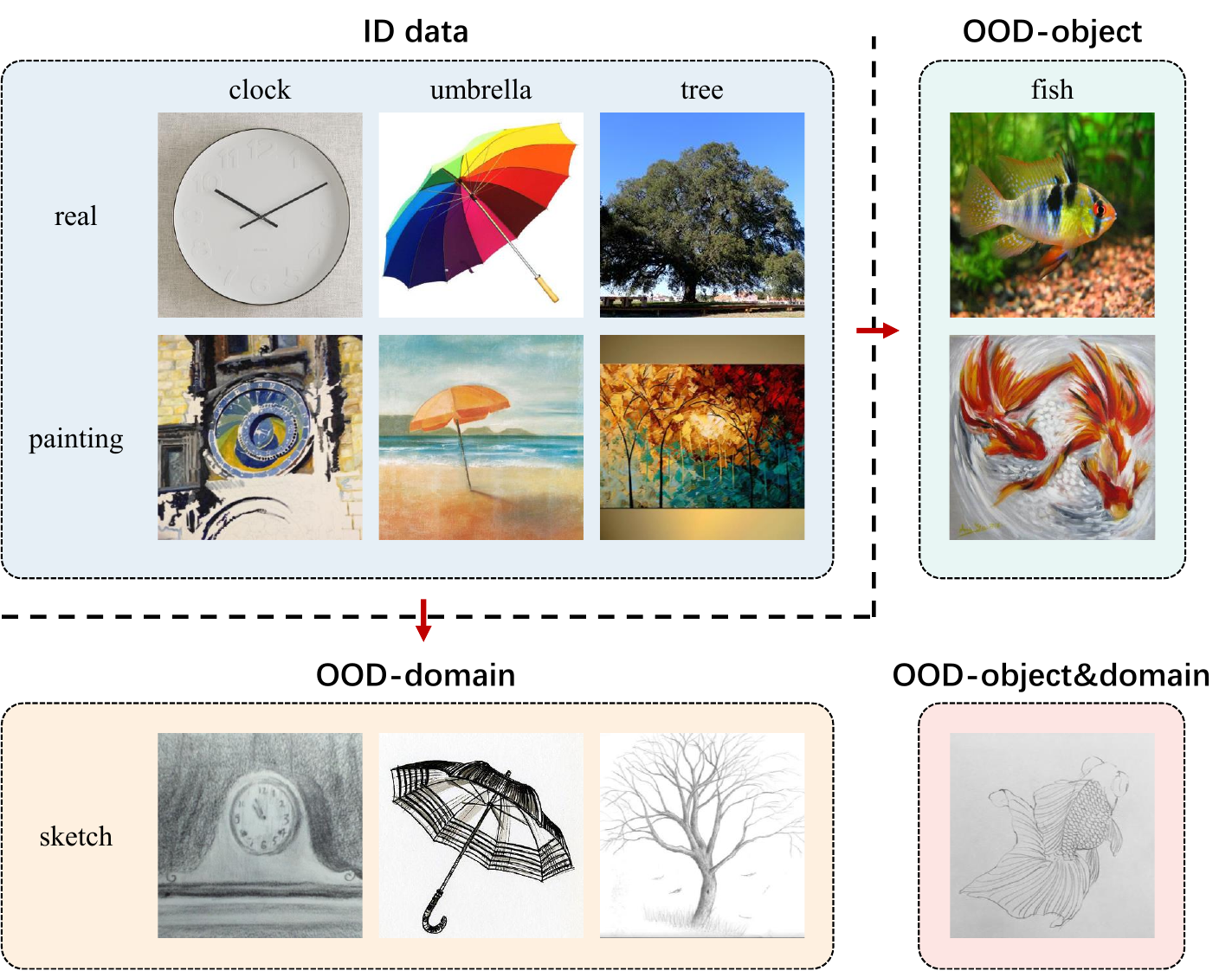}
\caption{An illustration of the partition of the DomainNet OOD dataset. Using the same ID data as the training set, OOD data is partitioned separately along the domain and object perspectives for testing purposes.}
\label{fig:domainnet}
\end{figure}

\subsection{DomainNet OOD Dataset}
This dataset is construct to study the impact of categorization perspectives in ID data on a model's OOD detection performance.
This requires each sample in the dataset to include multiple perspectives of labels. We find that DomainNet \cite{DomainNet_2019}, commonly used for domain generalization tasks, provides both domain and object labels for each data sample. Therefore, we choose to construct the target OOD detection dataset based on it.

DomainNet consists of images from 6 domains and 350 object classes in total. We randomly select 5 domains as ID domains and 300 object classes as ID objects, treating the remaining domain and objects as the OOD domain and objects, respectively, as illustrated in \cref{fig:domainnet}.

In the ``object-separated'' task, the model is trained to perform classification across the 300 ID object classes. In contrast, in the ``domain-separated'' task, the model is trained to classify across the 5 ID domains. By modifying the categorization perspective of the ID data, we make the model learn different Semantic Difference Spaces.

We then design two testing protocols, both utilizing the test set of the 300 ID objects within the 5 ID domains as the ID data. In the ``OOD-object'' protocol, OOD data are drawn from the remaining 50 OOD object classes while ensuring their domains belong to the 5 ID domains. This maintains the consistency in the domain distribution between OOD and ID data, allowing us to specifically evaluate the model's OOD detection capability in the object perspective.
In the ``OOD-domain'' protocol, we select OOD data from the remaining OOD domain while ensuring the objects belong to the ID object classes. This setup enables an evaluation of the model's OOD detection capability along the domain perspective.

\begin{figure}[t]%
\centering
\includegraphics[width=0.7\textwidth]{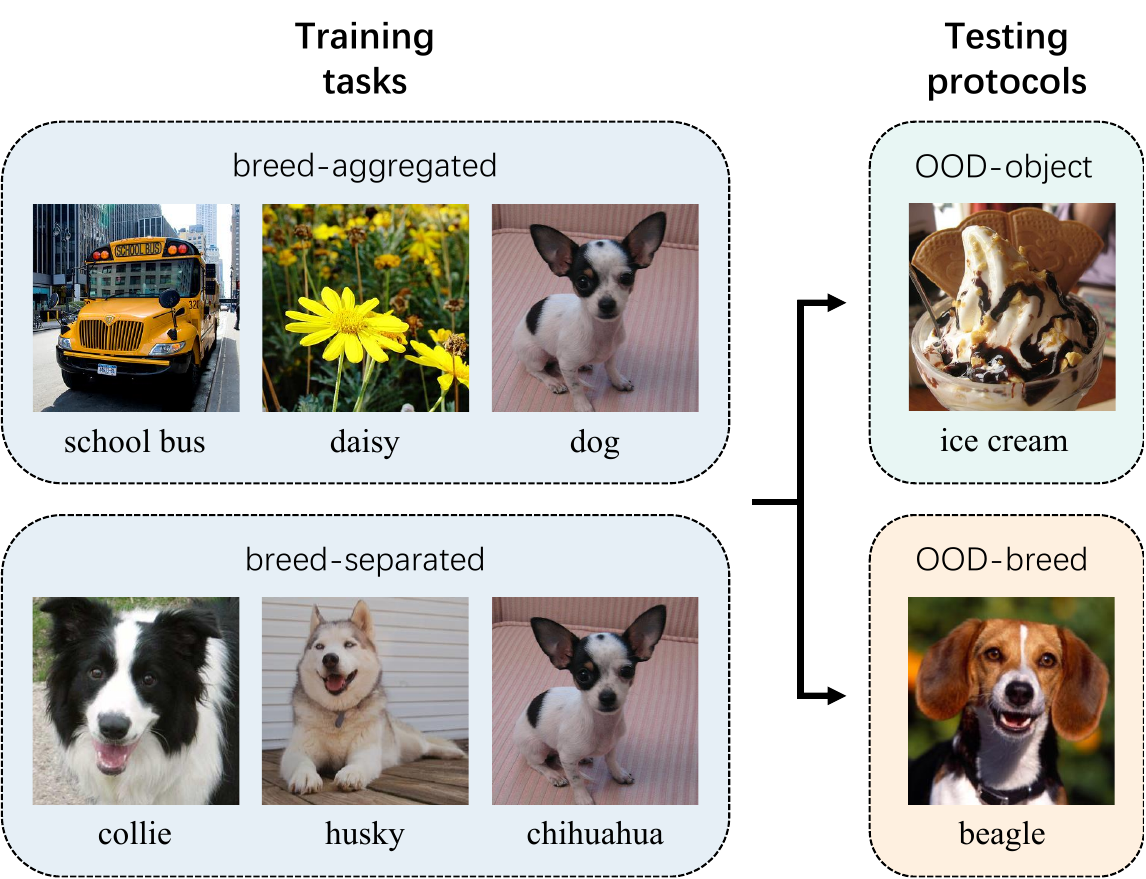}
\caption{An illustration of the ImageNet Dogs Dataset. It depicts two training tasks and two testing protocols.}
\label{fig:intro_IN1Kdog}
\end{figure}

\subsection{ImageNet Dogs Dataset}
We also aim to investigate the impact of classification granularity on a model's OOD detection performance.
Notably, ImageNet-1K \cite{ImageNet_2009} contains a substantial number of dog-related classes (126 classes representing various dog breeds). Thus, we utilize the classes in this ``dog subset'' to alter the classification granularity of the ID data.

The two training tasks are illustrated in \cref{fig:intro_IN1Kdog}. We first randomly select 100 classes from the ``dog subset''. In the ``breed-aggregated'' task, we aggregate the 100 selected dog classes into a single class and then randomly select 99 additional non-dog classes from ImageNet-1K, resulting in a total of 100 classes for training. As a comparison, in the ``breed-separated'' task, we take the selected dog classes as 100 different classes for training. These two settings, with varying classification granularities, result in different Semantic Difference Spaces learned by the model.

During testing, we adopt two distinct testing protocols. In the ``OOD-object'' protocol, 20 other non-dog classes from ImageNet-1K (excluding the selected 99 non-dog ID classes) are randomly selected as OOD data. This protocol is designed to assess the object-level OOD detection performance of a model. In the ``OOD-breed'' protocol, we use the test set of the 100 selected dog classes as ID data and the test set of the remaining 26 dog classes as OOD data. This testing protocol evaluates the OOD detection model's ability to identify novel dog classes under the two training tasks.

\section{Experiments}
\label{experiments}

\subsection{Experimental Details}
In the experiments on the Synthetic Dataset, we sample 1000 data from each of the 4 ID classes as the training data, with $\sigma=2$ for the class centers defined in Eq.~(\ref{eq:exp_vectors}). We use a linear classifier with a weight matrix $\boldsymbol{W}\in \mathbb{R}^{k\times d}$ as the model, where $k=4$ denotes the number of classes and $d=4$ represents the input dimension. We then train the classifier for 1000 epochs, using a SGD optimizer with a learning rate of 0.01, momentum of 0.9, and weight decay of 0.01.

For the rest of the experiments on real-world image datasets, a ResNet-50 \cite{ResNet_2016} is used as the backbone for the classification model, which is randomly initialized and trained on the ID data within each OOD settings. We follow the recipe provided by the official PyTorch guidelines for training on ImageNet-1K. The batch size is set to 32, with 90 training epochs. The model is optimized using the SGD optimizer with an initial learning rate of 0.1, momentum of 0.9, and weight decay of 1e-4. A Step Learning Rate Scheduler is employed, with a step size of 30 epochs and a gamma value of 0.1.

In the OSP experiment, the linear probe is trained independently using 50\% of the OOD dataset and an equal number of randomly sampled ID (ImageNet-1K) data. The probe is optimized with the Adam optimizer (learning rate 1e-3, weight decay 0.01) for 10 epochs. The total OSP computation time across one evaluated setting is approximately 10 minutes on a single GPU. The sensitivity of OSP scores to the subset size of OOD data used for training is analyzed in Appendix C. To evaluate whether the framework generalizes beyond ResNet-50, we additionally conduct the OSP assessment using ViT-B/16 as the backbone; results and analysis are provided in Appendix B.

In all the experiments, we include a total of 10 post-hoc methods: MSP \cite{MSP_2017}, ODIN \cite{ODIN_2018}, EBO \cite{EBO_2020}, GradNorm \cite{GradNorm_2021}, RMDS \cite{RMDS_2021}, KNN \cite{KNN_2022}, DICE \cite{DICE_2022}, ASH \cite{ASH_2023}, Relation \cite{Relation_2023}, and SCALE \cite{SCALE_2024}, as well as 5 training-based methods: GODIN \cite{GODIN_2020}, LogitNorm \cite{LogitNorm_2022}, VOS \cite{VOS_2022}, CIDER \cite{CIDER_2023}, and ISH \cite{SCALE_2024}.
The implementation of these OOD detection methods follows the code provided in the OpenOOD \cite{OpenOOD_2022} repository. Among these methods, post-hoc approaches directly use the pre-trained model for testing, while training-based approaches require further fine-tuning on the training data before testing. 

The widely used AUROC is employed in all the experiments as the metric to evaluate the performance of the OOD detection methods. To guarantee robust results, all classification models in this section are trained over three independent runs using different random seeds. We report the mean ± standard deviation in the tables.

\begin{table*}[t]
\centering
\caption{AUROC (\%) of OOD detection methods on Synthetic Dataset}\label{tab:exp_demo_1}
\resizebox{1.0\textwidth}{!}{
\renewcommand{\arraystretch}{1.5}
\begin{tabular}{c|c|c|c|c|c|c}
\toprule
                              & \multicolumn{3}{c|}{\textbf{w/o shift in Semantic Difference Space}}                              & \multicolumn{3}{c}{\textbf{with shift in Semantic Difference Space}}                             \\
\midrule
$\boldsymbol{s}_o$            & \multicolumn{3}{c|}{$[\sigma,\sigma,0,0]^\top$}                                          & \multicolumn{3}{c}{$[0,\sigma,0,0]^\top$}                                               \\
\midrule
$\boldsymbol{c}_o$            & $[0,0,\sigma,\sigma]^\top$ & $[0,0,-\sigma,\sigma]^\top$ & $[0,0,-\sigma,-\sigma]^\top$ & $[0,0,\sigma,\sigma]^\top$ & $[0,0,-\sigma,\sigma]^\top$ & $[0,0,-\sigma,-\sigma]^\top$ \\
\midrule
MSP~\cite{MSP_2017}           & 51.6±1.2 & 51.4±0.8 & 50.5±0.9 & 78.4±0.8 & 78.6±0.9 & 79.5±0.4 \\
EBO~\cite{EBO_2020}           & 49.8±0.2 & 50.2±1.3 & 50.5±0.6 & 75.2±1.4 & 73.2±0.7 & 73.7±1.0 \\
GradNorm~\cite{GradNorm_2021} & 49.5±0.5 & 50.6±1.0 & 50.4±0.8 & 73.8±0.4 & 74.7±1.1 & 74.2±0.3 \\
\bottomrule
\end{tabular}
}
\end{table*}

\subsection{Results on Synthetic Data}

Since the input vectors are directly provided as ground truth features, we exclude OOD detection methods that might involve manipulation of the feature space to avoid potential ``cheating''.
The results of three tested methods are shown in \cref{tab:exp_demo_1}. When the OOD class center does not exhibit shifts in the Semantic Difference Space (the three left columns of the results), all the OOD detection methods fail to distinguish between the OOD and ID distributions, with detection AUROC around 50\%, regardless of whether there is a shift in the Covariate Space. This indicates that the OOD detection task becomes intractable for these methods.
In contrast, when the OOD class center includes a shift in the Semantic Difference Space (the three right columns of the results), the OOD detection methods can effectively identify the OOD samples, achieving high AUROC scores. 

Additionally, both sets of experimental results clearly show that the changes in the Covariate Space of the OOD class center do not influence the performance of the OOD detection methods. The findings support the conclusion in Theorem 1, which asserts that when the difference between class centers is $\mathbf{0}$ in the Semantic Difference Space, the OOD detection model will yield indistinguishable outputs for them.

\subsection{Results on DomainNet OOD}

\begin{table*}[t]
\centering
\caption{AUROC (\%) of OOD detection methods on DomainNet OOD with Quickdraw as OOD domain}\label{tab:exp_DN_OOD}
\resizebox{0.6\textwidth}{!}{
\renewcommand{\arraystretch}{1.2}
\begin{tabular}{c|c|c|c|c}
\toprule
Testing Protocol                & \multicolumn{2}{c|}{OOD-object} & \multicolumn{2}{c}{OOD-domain} \\
\midrule
Training Perspective              & object & domain & object & domain       \\
\midrule
\multicolumn{5}{c}{Post-hoc Method} \\
\midrule
MSP~\cite{MSP_2017}             & 75.1±0.4 & 51.4±0.2 & 80.4±0.1 & 94.8±0.4  \\
ODIN~\cite{ODIN_2018}           & 73.2±0.4 & 51.2±0.3 & 93.8±0.3 & 96.0±1.3  \\
EBO~\cite{EBO_2020}             & 74.2±0.6 & 51.7±0.3 & 86.4±0.2 & 88.8±1.8  \\
GradNorm~\cite{GradNorm_2021}   & 63.1±0.5 & 52.4±0.5 & 80.8±2.9 & 97.2±0.4  \\
RMDS~\cite{RMDS_2021}           & 72.8±0.4 & 50.5±0.1 & 75.1±1.3 & 89.5±1.2  \\
KNN~\cite{KNN_2022}             & 65.2±0.3 & 51.3±0.3 & 88.0±0.4 & 53.9±18.9 \\
DICE~\cite{DICE_2022}           & 69.3±0.3 & 51.4±0.3 & 86.8±0.9 & 96.5±2.2  \\
ASH~\cite{ASH_2023}             & 72.1±0.4 & 51.7±0.5 & 84.0±1.1 & 95.4±0.6  \\
Relation~\cite{Relation_2023}   & 66.3±0.4 & 49.4±0.2 & 79.6±0.2 & 88.1±3.7  \\
SCALE~\cite{SCALE_2024}         & 73.1±0.3 & 51.7±0.7 & 84.6±1.4 & 96.9±0.7  \\
\midrule
\multicolumn{5}{c}{Training-based Method} \\
\midrule
GODIN~\cite{GODIN_2020}         & 71.6±0.3 & 49.3±0.2 & 93.6±0.6 & 94.6±2.4  \\
LogitNorm~\cite{LogitNorm_2022} & 75.8±0.4 & 52.2±0.2 & 83.8±0.4 & 91.3±1.9  \\
VOS~\cite{VOS_2022}             & 74.5±0.5 & 52.2±0.1 & 86.9±0.3 & 89.5±1.7  \\
CIDER~\cite{CIDER_2023}         & 63.9±0.3 & 51.1±0.3 & 88.1±0.6 & 53.5±18.4 \\
ISH~\cite{SCALE_2024}           & 75.0±0.4 & 51.9±0.1 & 80.4±0.4 & 94.3±0.7 \\
\bottomrule
\end{tabular}
}
\end{table*}

\textbf{OOD Detection Performance.}
The experiments on DomainNet OOD are conducted to examine how detection results are influenced by using ID data categorized from different perspectives. As shown in \cref{tab:exp_DN_OOD}, the model's OOD detection performance undergoes notable variations depending on the categorization perspectives in the training tasks. When the partition perspective of the OOD samples matches the training categorization perspective (e.g., the testing protocol is to detect novel objects, denoted as ``OOD-object'', and the training task is also to classify across different objects, denoted as ``object''), most OOD detection methods are effective to some extent, with an average AUROC of around 70\%. However, when the OOD shift does not match the training perspective (e.g., keep the testing protocol as ``OOD-object'' while modifying the training perspective to classify across different domains, denoted as ``domain''), all the model's OOD detection performance significantly decreases, with an average AUROC of around 50\%.

Additionally, we observe an anomaly: nearest neighbor-based methods like KNN~\cite{KNN_2022} exhibit highly unstable and degraded performance in the ``OOD-domain / domain'' setting (53.9$\pm$18.9\%), with AUROC fluctuating dramatically around random chance across runs. This occurs because, under the domain-separated training task, the feature space is structured to separate domains rather than objects. OOD domain samples may thus share local feature neighborhoods with certain ID domain samples, causing KNN to misclassify them as ID. A similar pattern is observed for CIDER~\cite{CIDER_2023} (53.5$\pm$18.4\%), which also relies on feature-space distances. This highlights a limitation of distance-based methods in domain-structured feature spaces.

\textbf{Results for Different Domain Partitions.}
In this part, we study the impact of different domain partitions. When Quickdraw serves as the OOD domain, as in \cref{tab:exp_DN_OOD}, we find that all OOD detection methods remain highly effective (with an average AUROC of over 70\%) in the ``OOD-domain'' protocol even under an unmatched setting (e.g., with the ``object'' training perspective). We thus explore how partitioning different OOD domains influences the model's detection of the novel domain. Specifically, we randomly re-partition the ID and OOD objects and domains, and conduct two additional experiments. In these two experiments, Infograph and Clipart are randomly selected as OOD domains, and their experimental results are shown in \cref{tab:exp_DN_OOD_infograph} and \cref{tab:exp_DN_OOD_clipart}.

\begin{table*}[t]
\centering
\caption{AUROC (\%) of OOD detection methods on DomainNet OOD with Infograph as OOD domain}\label{tab:exp_DN_OOD_infograph}
\resizebox{0.6\textwidth}{!}{
\renewcommand{\arraystretch}{1.2}
\begin{tabular}{c|c|c|c|c}
\toprule
Testing Protocol                & \multicolumn{2}{c|}{OOD-object} & \multicolumn{2}{c}{OOD-domain} \\
\midrule
Training Perspective              & object & domain & object & domain       \\
\midrule
MSP~\cite{MSP_2017}           & 76.3±0.3 & 51.7±0.3 & 76.2±0.4 & 82.4±1.0 \\
ODIN~\cite{ODIN_2018}         & 67.4±0.4 & 50.5±0.4 & 63.7±1.1 & 50.9±0.9 \\
EBO~\cite{EBO_2020}           & 75.7±0.3 & 52.0±0.2 & 79.5±0.7 & 83.6±0.9 \\
GradNorm~\cite{GradNorm_2021} & 63.2±0.1 & 52.5±0.5 & 48.1±2.1 & 73.3±2.5 \\
RMDS~\cite{RMDS_2021}         & 74.5±0.3 & 50.4±0.2 & 77.2±0.8 & 73.4±3.1 \\
KNN~\cite{KNN_2022}           & 66.8±0.1 & 51.1±0.3 & 82.7±0.3 & 90.3±0.2 \\
DICE~\cite{DICE_2022}         & 69.5±0.3 & 52.1±0.1 & 61.1±1.0 & 74.7±1.3 \\
ASH~\cite{ASH_2023}           & 73.3±0.2 & 52.1±0.2 & 72.5±1.4 & 80.7±3.5 \\
Relation~\cite{Relation_2023} & 68.0±0.3 & 51.2±0.0 & 73.8±0.4 & 68.2±1.2 \\
SCALE~\cite{SCALE_2024}       & 73.9±0.5 & 52.2±0.8 & 71.5±2.5 & 81.2±1.9 \\
\bottomrule
\end{tabular}
}
\end{table*}

\begin{table*}[t]
\centering
\caption{AUROC (\%) of OOD detection methods on DomainNet OOD with Clipart as OOD domain}\label{tab:exp_DN_OOD_clipart}
\resizebox{0.6\textwidth}{!}{
\renewcommand{\arraystretch}{1.2}
\begin{tabular}{c|c|c|c|c}
\toprule
Testing Protocol                & \multicolumn{2}{c|}{OOD-object} & \multicolumn{2}{c}{OOD-domain} \\
\midrule
Training Perspective              & object & domain & object & domain       \\
\midrule
MSP~\cite{MSP_2017}           & 73.1±0.2 & 51.0±0.1 & 53.0±0.1 & 84.5±0.2  \\
ODIN~\cite{ODIN_2018}         & 64.8±1.4 & 52.0±0.1 & 47.0±1.8 & 60.3±3.6  \\
EBO~\cite{EBO_2020}           & 71.9±0.3 & 51.1±0.1 & 56.5±0.3 & 85.2±0.2  \\
GradNorm~\cite{GradNorm_2021} & 60.7±0.2 & 51.3±0.5 & 37.4±1.1 & 78.2±1.0  \\
RMDS~\cite{RMDS_2021}         & 71.6±0.3 & 50.7±0.1 & 58.8±0.7 & 71.0±10.9 \\
KNN~\cite{KNN_2022}           & 64.5±0.3 & 50.7±0.1 & 73.7±0.3 & 83.9±0.9  \\
DICE~\cite{DICE_2022}         & 67.0±0.5 & 51.0±0.1 & 46.2±1.0 & 80.5±2.3  \\
ASH~\cite{ASH_2023}           & 70.5±0.4 & 51.1±0.0 & 47.6±0.9 & 83.8±0.4  \\
Relation~\cite{Relation_2023} & 68.3±0.1 & 49.5±0.1 & 54.6±0.1 & 69.9±0.7  \\
SCALE~\cite{SCALE_2024}       & 70.7±0.2 & 51.1±0.3 & 47.2±1.3 & 82.0±1.2 \\
\bottomrule
\end{tabular}
}
\end{table*}

Similar to the results in \cref{tab:exp_DN_OOD}, when using the ``OOD-domain'' testing protocol, the performance of OOD detection methods is worse under the ``object'' training perspective than under the ``domain'' training perspective. However, we find the model's ability to detect the novel domain under the ``object'' training perspective varies a lot across different domain partitions. The overall OOD detection performance of the evaluated methods is highest under the Quickdraw experiment (in \cref{tab:exp_DN_OOD}) and lowest under the Clipart experiment (in \cref{tab:exp_DN_OOD_clipart}).

We further investigate the reasons behind the changes in the model's ability to detect OOD domains in the three domain partition experiments. We test the model's object classification performance on both ID and OOD domains in these three cases, as shown in \cref{tab:exp_DN_OOD_ACC}.

\begin{table*}[t]
\centering
\caption{Accuracy (\%) on ID and OOD domains for the classification model trained using the object-separated setup}\label{tab:exp_DN_OOD_ACC}
\resizebox{0.6\textwidth}{!}{
\renewcommand{\arraystretch}{1.2}
\begin{tabular}{c|c|c|c|c|c|c}
\toprule
OOD   Domain & \multicolumn{2}{c|}{\textbf{Quickdraw}} & \multicolumn{2}{c|}{\textbf{Infograph}} & \multicolumn{2}{c}{\textbf{Clipart}} \\
\midrule
Split        & ID                 & OOD               & ID                 & OOD               & ID                & OOD              \\
\midrule
Top-1 Acc            & 69.4±0.6 & 14.5±0.9 & 73.9±0.3 & 21.9±0.3 & 70.8±0.3 & 61.7±0.6 \\
Top-5 Acc            & 85.1±0.5 & 31.5±1.5 & 90.4±0.2 & 35.6±0.3 & 87.4±0.2 & 80.4±0.6 \\
\bottomrule
\end{tabular}
}
\end{table*}

It can be observed that when Quickdraw is used as the OOD domain, the model achieves the lowest classification accuracy on the OOD domain, whereas when Clipart is used as the OOD domain, the model achieves the highest classification accuracy on the OOD domain (likely because Clipart uses real-world color, shading, and detail at a similar visual fidelity to many ID domains, particularly ``painting'' and ``real''). Interestingly, in contrast to the classification performance, the model achieves the highest overall OOD detection performance when Quickdraw is the OOD domain (as shown in \cref{tab:exp_DN_OOD}), and performs the worst when Clipart is the OOD domain (as shown in \cref{tab:exp_DN_OOD_clipart}). The above results suggest that both the model's OOD generalization and detection capabilities are determined by the shift within the model's ``Semantic Difference Space'' between the OOD and ID domains. Taking the Quickdraw setting as an example, the shift between Quickdraw and other ID domains is relatively large, leading to the best OOD detection performance. However, such a shift severely impacts the model's cross-domain generalization, resulting in the worst cross-domain classification performance.

\textbf{Important Conclusions.}
Based on the experiments conducted on the DomainNet OOD dataset, we can draw the following conclusions: 
\begin{itemize}
    \item Within the same testing protocol, the model trained under the matched perspective (the categorization perspective of training data is the same as the partition perspective of ID and OOD data) exhibits better OOD detection capability than the model trained under the unmatched perspective.    
    \item To some extent, the model's detection and generalization abilities on OOD samples are in conflict within the same OOD setting. Detecting OOD samples requires a significant difference within the model's ``Semantic Difference Space'' between OOD and ID data, but this affects the model's generalization performance.
\end{itemize}

\subsection{Results on ImageNet Dogs}

\begin{table*}[t]
\centering
\caption{AUROC (\%) of OOD detection methods on ImageNet Dogs}\label{tab:exp_IN_dog}
\resizebox{0.6\textwidth}{!}{
\renewcommand{\arraystretch}{1.2}
\begin{tabular}{c|c|c|c|c}
\toprule
Testing Protocol                & \multicolumn{2}{c|}{OOD-object} & \multicolumn{2}{c}{OOD-breed} \\
\midrule
Training Granularity              & aggregated & separated       & aggregated & separated      \\
\midrule
\multicolumn{5}{c}{Post-hoc Method} \\
\midrule
MSP~\cite{MSP_2017}             & 81.1±0.5 & 95.2±0.5 & 51.4±0.6 & 74.0±0.2 \\
ODIN~\cite{ODIN_2018}           & 80.5±0.7 & 96.5±0.3 & 52.6±0.6 & 75.4±0.2 \\
EBO~\cite{EBO_2020}             & 80.7±0.8 & 97.8±0.2 & 52.2±0.8 & 76.2±0.3 \\
GradNorm~\cite{GradNorm_2021}   & 61.5±0.8 & 94.3±1.0 & 50.1±0.6 & 58.9±2.0 \\
RMDS~\cite{RMDS_2021}           & 80.3±0.4 & 93.5±0.6 & 50.7±0.5 & 76.3±0.3 \\
KNN~\cite{KNN_2022}             & 74.7±0.6 & 98.0±0.2 & 50.3±0.5 & 75.4±0.8 \\
DICE~\cite{DICE_2022}           & 76.5±0.9 & 97.3±0.5 & 51.0±0.6 & 71.5±0.7 \\
ASH~\cite{ASH_2023}             & 76.6±1.5 & 97.2±0.4 & 52.1±1.3 & 69.3±1.8 \\
Relation~\cite{Relation_2023}   & 79.5±0.6 & 97.0±0.2 & 52.1±1.3 & 73.8±0.6 \\
SCALE~\cite{SCALE_2024}         & 78.2±0.5 & 97.8±0.2 & 51.5±0.6 & 74.3±1.3 \\
\midrule
\multicolumn{5}{c}{Training-based Method} \\
\midrule
GODIN~\cite{GODIN_2020}         & 80.0±1.2 & 93.4±0.9 & 53.0±1.0 & 75.2±0.5 \\
LogitNorm~\cite{LogitNorm_2022} & 81.7±0.4 & 97.7±0.1 & 51.2±0.7 & 76.5±0.3 \\
VOS~\cite{VOS_2022}             & 80.7±0.6 & 97.6±0.2 & 52.2±0.8 & 76.2±0.4 \\
CIDER~\cite{CIDER_2023}         & 74.1±0.5 & 98.0±0.2 & 50.0±0.3 & 74.0±0.6 \\
ISH~\cite{SCALE_2024}           & 81.1±0.3 & 94.8±0.4 & 51.1±0.7 & 73.9±0.1 \\
\bottomrule
\end{tabular}
}
\end{table*}

\textbf{OOD Detection Performance.}
We next conduct experiments on ImageNet Dogs to investigate how the classification granularity within the ID data influences the model's OOD detection performance, as shown in \cref{tab:exp_IN_dog}. \revise{(Additional results from more repeated experiments are provided in Appendix F.)}

For the ``OOD-object'' testing protocol, aimed at detecting OOD non-dog classes, we can see the AUROCs of OOD detection methods approach 80\% when using the ``aggregated'' training granularity (where the model is trained using an aggregated ``dog'' class and 99 non-dog classes), indicating that the model learns enough information in the ``Semantic Difference Space'' to distinguish the object-level OOD samples. Surprisingly, the AUROCs of most methods are higher and exceed 95\% when using the ``separated'' training granularity (where the model is trained using an 100 separated breed-level classes). We believe this occurs because, although the model only learns breed-related information within the ``Semantic Difference Space'' in this case, the object-level OOD samples, which contain no breed-related information at all, still can be easily distinguished from the ID data. This explains why detection methods all perform quite well on the so-called ``Far-OOD'' tasks in existing benchmarks.

As for the ``OOD-breed'' testing protocol, the OOD targets consist of novel breed classes. When the training granularity is ``aggregated'', the coarse-grained object-level training classification granularity does not match the fine-grained breed-level partition granularity of OOD and ID data in the testing protocol. As a result, the AUROCs of all OOD detection methods on the ``OOD-breed'' setting drop to around 50\%, indicating that these methods struggle to differentiate between OOD and ID dog classes.
In contrast, when the training granularity is ``separated'', most OOD detection methods achieve an AUROC of over 70\% on the ``OOD-breed'' testing protocol. This indicates that the model learns breed-related features in this settings and is thus capable of distinguishing the novel dog classes from the ID dog classes.

\begin{figure}[t]%
\centering
\includegraphics[width=0.7\textwidth]{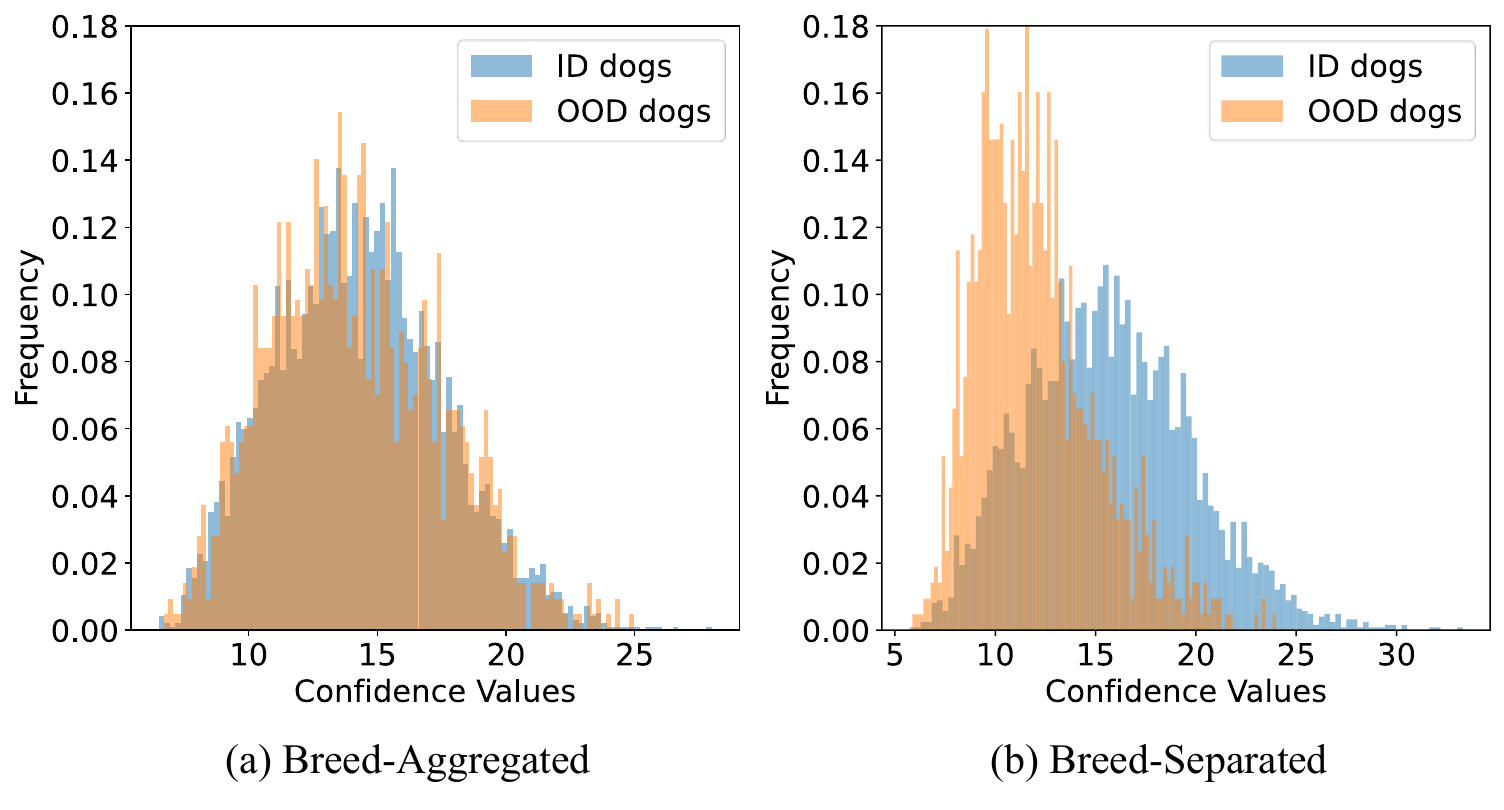}
\caption{The distributions of confidence output under two training tasks (EBO~\cite{EBO_2020} shown as a representative example; other methods exhibit similar patterns)}. In the ``breed-aggregated'' training task, the two confidence distributions are almost overlapping, while in the ``breed-separated'' training task, the confidence distributions for ID and OOD dogs show a significant difference.\label{fig:exp_conf_value}
\end{figure}

\textbf{Distributions of confidence values.} 
We present the distribution of confidence values obtained by EBO \cite{EBO_2020} across different training tasks for ID and OOD samples in the ``OOD-breed'' testing protocol, as shown in \cref{fig:exp_conf_value}. When the training task is ``breed-aggregated'', the distributions of confidence values for ID and OOD dog classes are nearly identical.
In contrast, under the ``breed-separated'' training task, there is a significant difference in the distribution of confidence values between novel dog classes and ID dog classes, allowing the model to effectively distinguish between them.

\begin{figure}[t]%
\centering
\includegraphics[width=0.7\textwidth]{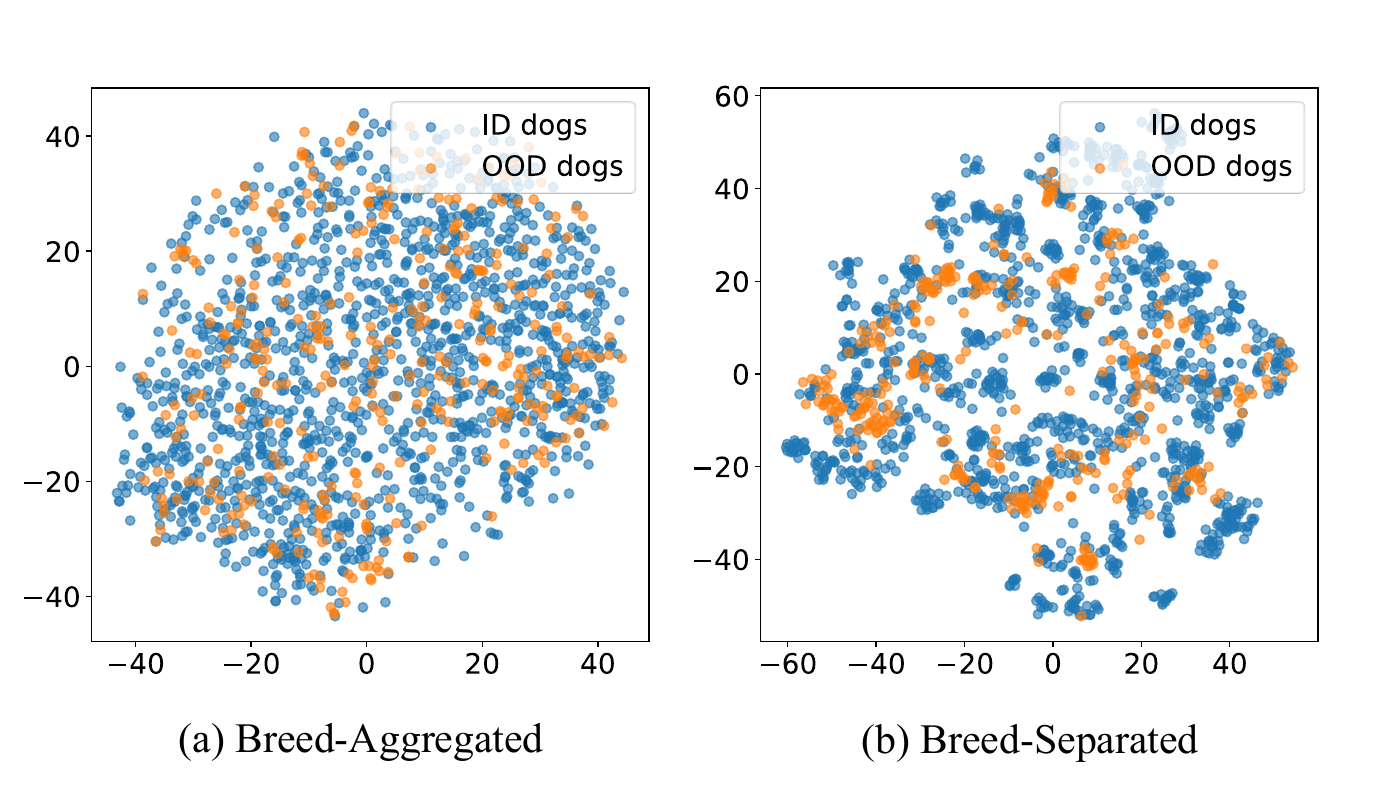}
\caption{The t-SNE visualization of the features of ID and OOD dogs under ``breed-aggregated'' and ``breed-separated'' tasks. The feature distributions of OOD and ID dogs are highly similar in the ``breed-aggregated'' task but exhibit clear distinctions in the ``breed-separated'' task.}
\label{fig:exp_tsne}
\end{figure}

\textbf{Feature visualization.}
Finally, we visualize the features of ID and OOD dog classes learned by the model in both the ``breed-separated'' and ``breed-aggregated'' tasks using t-SNE, as shown in \cref{fig:exp_tsne}. 
In the ``breed-aggregated'' task, the features of all dog classes cluster together with no clear separation. This lack of differentiation makes it hard for models to detect novel dog classes as OOD samples.
However, in the ``breed-separated'' task, features of different ID dog classes are separated, with most OOD dog classes positioned in the gap between the ID classes. This allows the OOD detection methods to effectively identify these ``outliers'' as OOD data.

\textbf{Important Conclusions.}
The above results on the constructed ImageNet Dogs dataset lead to the following conclusions:
\begin{itemize}
    \item After trained on ID data with fine-grained labels, a model can maintain the ability to detect coarser-grained OOD object (data with a completely different type from the ID data, e.g., ice cream vs. breed classification dataset). This is because these OOD samples contain no meaningful information within the fine-grained ``Semantic Difference Space'' and can be easily distinguished from the ID data. This explains why detection methods all perform quite well on the so-called ``Far-OOD'' tasks in existing benchmarks. 
    \item For the model trained with coarse-grained labels, detecting finer-grained OOD objects (such as a novel sub-class of the ID class, e.g., husky vs. dog) is challenging. Within the model's ``Semantic Difference Space'', these OOD samples are indistinguishable from their parent categories, causing the model to treat them as ID data.
\end{itemize}

\subsection{Assessment via OOD Separability}
\textbf{Assess Existing Benchmarks.}
In this experiment, we use the proposed OOD Separability Probe (OSP) method to measure the OOD Separability within several existing benchmarks and assess whether they are appropriate for OOD detection. Specifically, we evaluate several widely used OOD detection settings with ImageNet-1K as the ID data. The test datasets used as OOD data include SSB-Hard \cite{SSB-hard_2022}, NINCO \cite{NINCO_2023}, Textures \cite{Texture_2011}, iNaturalist \cite{MOS_iNaturalist_2021}, OpenImage-O \cite{ViM_OpenImage-O_2022}, ImageNet-V2 \cite{ImageNet-V2_2019}, ImageNet-C \cite{ImageNet-C_2019}, and ImageNet-R \cite{ImageNet-R_2021}. Among these, we intentionally include datasets exhibiting covariate shift, such as ImageNet-V2 and ImageNet-C, to investigate whether the measured OOD Separability remains predictive of detection performance for these datasets.

\begin{table*}[t]
\centering
\caption{AUROC (\%) of OOD detection methods and OOD Separability on the OOD datasets (with ImageNet-1K as the ID dataset)}\label{tab:exp_IN1K_full}
\resizebox{1.0\textwidth}{!}{
\renewcommand{\arraystretch}{1.2}
\begin{tabular}{c|c|c|c|c|c|c|c|c}
\toprule
         & \textbf{SSB-hard} & \textbf{NINCO} & \textbf{Textures} & \textbf{iNaturalist} & \textbf{OpenImage-O} & \textbf{ImageNet-V2} & \textbf{ImageNet-C} & \textbf{ImageNet-R} \\
\midrule
MSP~\cite{MSP_2017}           & 72.1 & 80.2 & 82.5 & 88.3 & 84.8 & 58.7 & 78.6 & 80.7 \\
ODIN~\cite{ODIN_2018}         & 71.8 & 77.4 & 88.7 & 90.9 & 88.1 & 57.8 & 77.0 & 85.7 \\
EBO~\cite{EBO_2020}           & 72.4 & 79.5 & 88.7 & 90.6 & 89.2 & 58.9 & 82.8 & 87.2 \\
GradNorm~\cite{GradNorm_2021} & 72.8 & 73.6 & 92.3 & 93.9 & 84.9 & 54.5 & 80.0 & 70.0 \\
RMDS~\cite{RMDS_2021}         & 71.5 & 82.5 & 85.9 & 87.2 & 85.7 & 58.8 & 79.9 & 81.9 \\
KNN~\cite{KNN_2022}           & 63.8 & 80.0 & 96.0 & 83.8 & 85.9 & 57.6 & 85.9 & 88.5 \\
DICE~\cite{DICE_2022}         & 70.8 & 75.8 & 92.0 & 92.5 & 88.5 & 57.3 & 82.5 & 83.1 \\
ASH~\cite{ASH_2023}           & 73.1 & 83.2 & 96.9 & 97.0 & 93.2 & 58.5 & 83.9 & 85.3 \\
Relation~\cite{Relation_2023} & 65.6 & 79.7 & 91.3 & 91.1 & 88.3 & 57.4 & 80.4 & 84.6 \\
SCALE~\cite{SCALE_2024}       & 77.2 & 85.2 & 97.5 & 97.9 & 93.8 & 58.6 & 86.2 & 84.0 \\
\midrule
$S(\mathcal{P}_I, \mathcal{P}_O)$      & 96.8 & 93.6 & 99.3 & 99.8 & 97.1 & 66.5 & 98.7 & 96.4  \\
\bottomrule
\end{tabular}
}
\end{table*}

The full experimental results, including the performance of commonly used OOD detection methods and the measured ``OOD Separability'' (denoted as $S(\mathcal{P}_I, \mathcal{P}_O)$), are presented in \cref{tab:exp_IN1K_full} (Note that a fixed public ImageNet-1K pre-trained checkpoint is used with no model training randomness). As observed, the performance gap across different OOD settings is more significant than that between various OOD detection methods, with an average AUROC of over 90\% on iNaturalist and less than 60\% on ImageNet-V2. In each OOD setting, the calculated OOD Separability $S(\mathcal{P}_I, \mathcal{P}_O)$ provides a feature-space reference for the detection performance of OOD detection methods and reflects how difficult the setting is under the representation learned by the evaluated backbone.

\begin{figure*}[t]%
\centering
\includegraphics[width=1.0\textwidth]{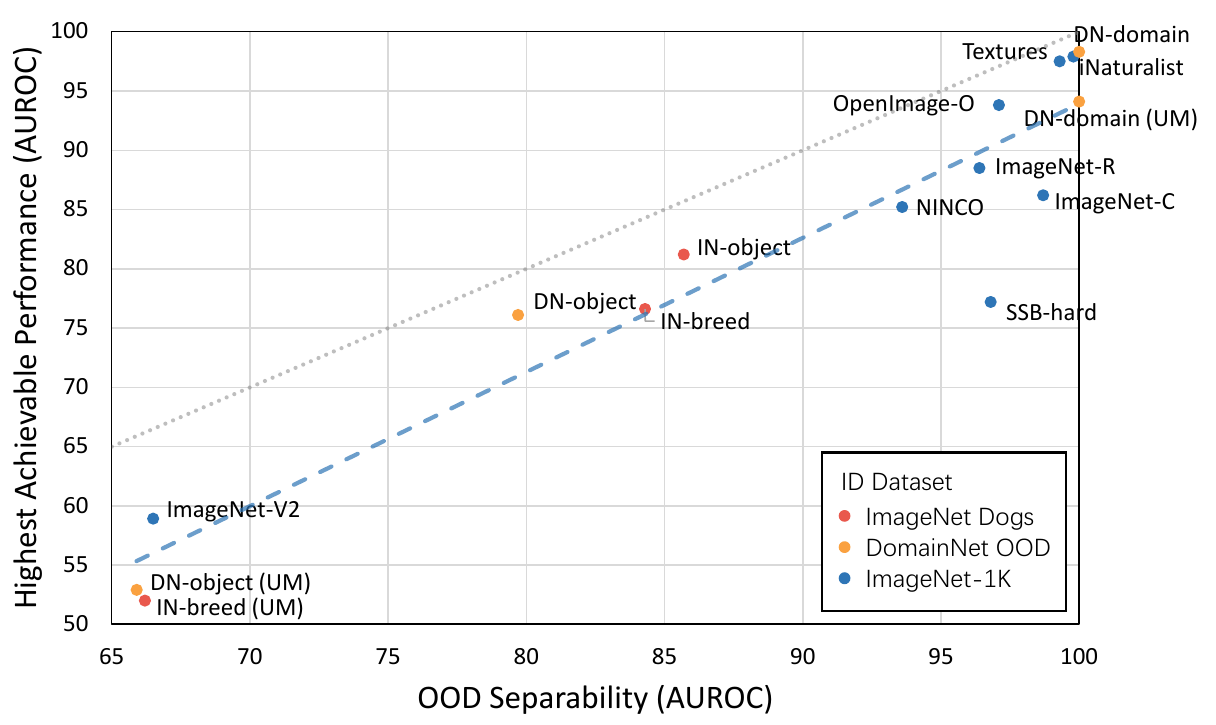}
\caption{The relationship between the OOD Separability and the highest achievable performance among the evaluated OOD detection methods. Each point represents a specific experimental setting, with the name of the OOD dataset labeled next to it. The ``UM'' in the name denotes a setting where the categories of the training data and the partition of OOD data are unmatched in perspective or granularity. The gray line indicates where detection performance equals OOD Separability; the blue dashed line is a linear fit across all settings.}\label{fig:exp_upper_bound}
\end{figure*}

We further plot a graph to illustrate the relationship between OOD Separability and the highest achievable performance for each OOD setting, as shown in \cref{fig:exp_upper_bound}. The OOD Separability is represented on the horizontal axis, while the highest performance achieved by the evaluated OOD detection methods is shown on the vertical axis. 

The gray line indicates the position where the OOD detection performance equals the OOD Separability. The blue dashed line represents the linear fit between the highest achievable performance and the OOD Separability across all the OOD settings. We can see that, overall, the best performance achievable in each OOD setting is positively correlated with the probe-measured OOD Separability. 

With the results shown in \cref{fig:exp_upper_bound}, we can draw the following important insights about the evaluated OOD settings and the corresponding benchmarks:
\begin{itemize}
    \item Both the OOD Separability and the detection performance on settings like ``iNaturalist'' and ``Textures'' are excessively high, indicating that these settings are readily handled by current OOD detection methods. While they remain valuable as sanity-check benchmarks, their near-saturated performance may provide limited resolution for comparing strong OOD detection methods.
    
    \item Certain settings like ``SSB-hard'' are far lower than the blue trend line. This suggests current OOD detection methods fail to capture the existing difference between OOD and ID data, which means it is still a highly worthwhile area of research. 

    \item Some testing datasets, such as ``ImageNet-C'' and ``ImageNet-R'', previously regarded as Covariate Shifted ID data, exhibit high OOD Separability as well as high OOD detection performance. This suggests that the training on ImageNet-1K may introduce unintended factors into the ``Semantic Difference Space'', providing clues for detecting these two datasets. More careful design is required during training if these data are not intended to be detected as OOD samples.
\end{itemize}

\begin{figure*}[t]%
\centering
\includegraphics[width=\textwidth]{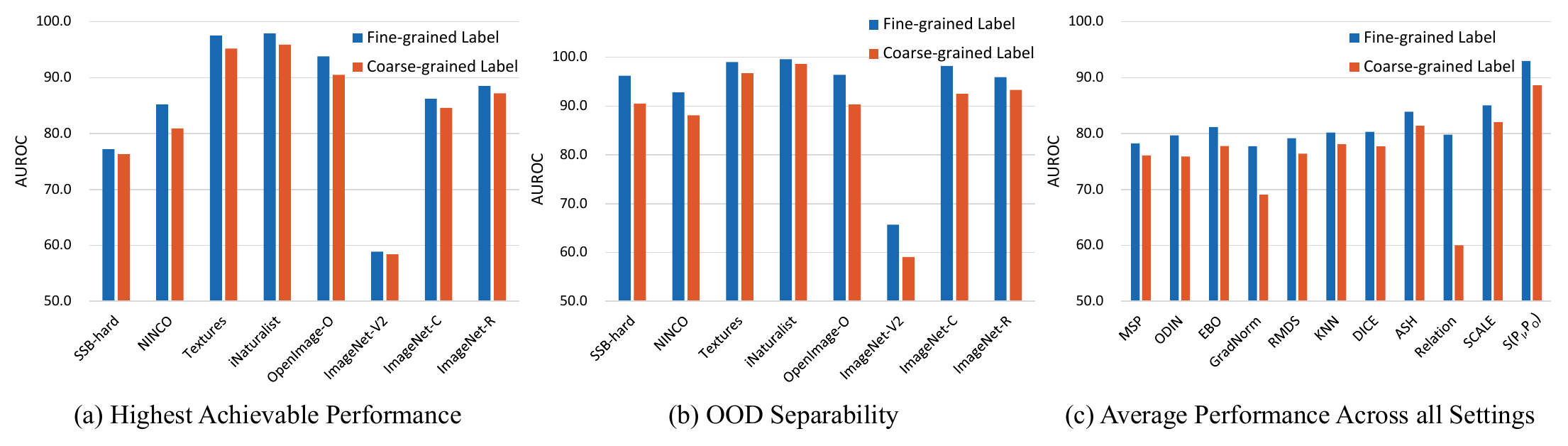}
\caption{Comparison of the model's OOD detection capability using ID data with different granularities. The detection capability experiences an overall decline when using ID data with ``Coarse-grained'' labels compared to using ``Fine-grained'' labels.}\label{fig:exp_label_level}
\end{figure*}

\textbf{Further Exploration on Granularity Effects.}
Finally, we further investigate the impact of the classification granularity of ID data using OOD Separability. Specifically, we treat the original ImageNet-1K labels as ``Fine-grained'' labels. Then, we use the semantic tree of ImageNet-1K to obtain the parent categories of each label, and we aggregate samples with the same parent category into one class to obtain ``Coarse-grained'' labels. We take the data with ``Coarse-grained'' labels to repeat the aforementioned OOD detection experiments. The comparison of results under both settings is shown in \cref{fig:exp_label_level}.

The results in \cref{fig:exp_label_level} can lead to the following conclusions:
\begin{itemize}
    \item After using ``Coarse-grained'' labels, both the highest achievable performance in each OOD setting and the corresponding OOD Separability decrease, as shown in \cref{fig:exp_label_level} (a) and (b). This indicates that models trained with ``Fine-grained'' labels consistently exhibit better OOD detection capabilities than the models trained with ``Coarse-grained'' labels.
    \item The impact of label granularity varies across detection methods, as shown in \cref{fig:exp_label_level}(c). Methods such as Relation \cite{Relation_2023} experience a significant performance drop when using the ``Coarse-grained'' labels, suggesting that these methods heavily rely on the detailed information provided by ``Fine-grained'' labels.
\end{itemize}

\section{Conclusion}
\label{conclusion}
In this paper, we introduce a formal definition of ``Semantic Difference Space'' to help address the inconsistent OOD partition criteria caused by the ambiguity of ``Semantic Shift'' in previous work. Through theoretical analysis, we demonstrate how the perspectives and granularities within ID distribution influence the model's OOD detection capability by affecting the range of the ``Semantic Difference Space''. Crucially, the proposed OOD Separability Probe (OSP) benefits the community by identifying benchmarks with high research value—specifically those where a significant gap exists between the current state-of-the-art performance and the OOD Separability of the benchmark. This insight directs future research toward more meaningful and challenging settings where OOD detection methods still have substantial room for improvement, rather than saturating on trivial tasks. However, our experiments are exclusively conducted on image classification, and the theoretical analysis involves simplifying assumptions that may not fully reflect the complexity of real-world deep learning systems. Future work should extend the experimental validation to other domains, such as object detection or text sentiment recognition, while refining the theoretical framework to better account for complex training dynamics.

\section*{Acknowledgments}
This work is partially supported by Strategic Priority Research Program of the Chinese Academy of Sciences (No. XDB0680202), Beijing Nova Program (20230484368), and Suzhou Frontier Technology Research Project (No. SYG202325).

\bibliographystyle{elsarticle} 
\bibliography{cas-refs}

\newpage
\appendix
\crefalias{section}{appendix}
\section{Proofs}
\label{app:proofs}

In this section, we will provide the proof for Proposition 1 and Theorem 1.

\subsection{Preliminary Definitions and Assumptions}

We begin by giving the following formal definition of the output prediction probabilities $p_i$ for each class of the linear classifier:
\begin{equation} \begin{aligned}
f(\boldsymbol{x}) &= softmax(\boldsymbol{W}\boldsymbol{x}) = [p_1(\boldsymbol{x}),...,p_k(\boldsymbol{x})]^\top, \\
p_i(\boldsymbol{x}) &= \frac{\exp{(\boldsymbol{W}_{i,:} \boldsymbol{x})}}{\sum_{j=1}^k \exp{(\boldsymbol{W}_{j,:} \boldsymbol{x})}}, i \in \{1,...,k\}.
\label{eq:softmax}
\end{aligned} \end{equation}

We then introduce the two mild assumptions for Proposition 1.

\textbf{Assumption 1.} \textit{Throughout the entire training process, the expected prediction probability of the linear classifier for each class is equal}
\begin{equation} \begin{aligned}
\mathbb{E}_{\boldsymbol{x}} [p_i(\boldsymbol{x})]=\frac{1}{k}, i \in \{1,...,k\}.
\label{eq:assumption_1}
\end{aligned} \end{equation}

Assumption 1 requires a balanced ID distribution across all classes, ensuring that the classification model does not develop any inherent bias toward a particular class during training.

\textbf{Assumption 2.} \textit{The sign of the covariance between the prediction probability for a specific class $i$ and the $j^{th}$ element of the input vector $\boldsymbol{x}$ matches that of its corresponding weight $\boldsymbol{W}_{ij}$}
\begin{equation} \begin{aligned}
\boldsymbol{W}_{ij} \cdot Cov(p_i(\boldsymbol{x}),\boldsymbol{x}_j) \geq 0, i \in \{1,...,k\}, j \in \{1,...,d\}.
\label{eq:assumption_2}
\end{aligned} \end{equation}

Assumption 2 is intuitively straightforward: when $\boldsymbol{W}_{ij}$ is positive, $p_i$ defined in \cref{eq:softmax} is a monotonically increasing function of $\boldsymbol{x}_j$, resulting in a covariance that is likely positive. Conversely, when $\boldsymbol{W}_{ij}$ is negative, the covariance is likely negative.
We propose Assumption 2 because calculating the analytical solution for the covariance is highly complex due to the involvement of the softmax function in $p_i$. Therefore, Assumption 2 is introduced for simplification.

\subsection{Necessary Lemma for Proposition 1}

To prove Proposition 1, we introduce the following lemma to ensure that the training process of the linear classifier remains unaffected by an orthogonal transformation.

\textbf{Lemma 1.} \textit{Given an orthogonal matrix $\boldsymbol{Q}$, training a linear classifier $f_1(\cdot)$ with a weight matrix $\boldsymbol{W}$ on the ID distribution $(\boldsymbol{x},y) \sim \mathcal{P}_I$ is equivalent to training a linear classifier $f_2(\cdot)$ with another weight matrix $\tilde{\boldsymbol{W}} = \boldsymbol{W} \cdot \boldsymbol{Q}^\top$ on the orthogonally transformed distribution $(\boldsymbol{x},y) \sim \tilde{\mathcal{P}}_I$. The orthogonally transformed distribution $\tilde{\mathcal{P}}_I$ denotes the distribution of $(\boldsymbol{Q} \boldsymbol{x},y)$ when $(\boldsymbol{x},y)$ is sampled from the original ID distribution $(\boldsymbol{x},y) \sim \mathcal{P}_I$.}

\textbf{Proof of Lemma 1.}
We only need to prove that if, at a certain training step $n$, the weight matrices of the two linear classifiers satisfy condition $\boldsymbol{W}_n = \tilde{\boldsymbol{W}}_n \cdot \boldsymbol{Q}$, and after their respective next optimization steps, the two weight matrices still satisfy $\boldsymbol{W}_{n+1} = \tilde{\boldsymbol{W}}_{n+1} \cdot \boldsymbol{Q}$, then it can be demonstrated that the two training setups are equivalent.

We can find that the outputs of the two models $f_1(\boldsymbol{x}_1)$ and $f_2(\boldsymbol{x}_2)$ are consistent when their inputs follows the relation $\boldsymbol{x}_2 = \boldsymbol{Q} \boldsymbol{x}_1$
\begin{equation} \begin{aligned}
f_1(\boldsymbol{x}_1) &= softmax(\boldsymbol{W}\boldsymbol{x}_1) \\
&= softmax((\tilde{\boldsymbol{W}} \cdot \boldsymbol{Q}) (\boldsymbol{Q}^\top \boldsymbol{x}_2)) \\
&= softmax(\tilde{\boldsymbol{W}} \boldsymbol{x}_2) \\
&= f_2(\boldsymbol{x}_2).
\end{aligned} \end{equation}

Since the training loss consists of cross-entropy loss and L2 regularization loss as follows (In the proofs of this lemma and subsequent Proposition 1, we assume that the training loss consists solely of these two components, consistent with the classification training in most works.)
\begin{equation} \begin{aligned}
L_1 &= - \mathbb{E}_{\boldsymbol{x},y \sim \mathcal{P}_I}[onehot(y)^\top \cdot log(f_1(\boldsymbol{x}))] + \frac{\lambda}{2} \Vert \boldsymbol{W} \Vert_2^2, \\
L_2 &= - \mathbb{E}_{\boldsymbol{x},y \sim \tilde{\mathcal{P}}_I}[onehot(y)^\top \cdot log(f_2(\boldsymbol{x}))] + \frac{\lambda}{2} \Vert \tilde{\boldsymbol{W}} \Vert_2^2,
\end{aligned} \end{equation}
where $L_1$ and $L_2$ denote the training loss of the two setups, and $onehot(y)$ refers to the one-hot vector of the label $y$.

We can then bridge the gradient of the two setups.
\begin{equation} \begin{aligned}
\frac{\partial L_1}{\partial \boldsymbol{W}} &= \mathbb{E}_{\boldsymbol{x},y \sim \mathcal{P}_I} [(f_1(\boldsymbol{x}) - onehot(y)) \boldsymbol{x}^\top] + \lambda \boldsymbol{W} \\
&= \mathbb{E}_{\boldsymbol{x},y \sim \mathcal{P}_I} [(f_2(\boldsymbol{Q} \boldsymbol{x}) - onehot(y)) \boldsymbol{x}^\top] + \lambda \tilde{\boldsymbol{W}} \cdot \boldsymbol{Q} \\
&= \mathbb{E}_{\boldsymbol{x},y \sim \mathcal{P}_I} [(f_2(\boldsymbol{Q} \boldsymbol{x}) - onehot(y)) \boldsymbol{x}^\top \boldsymbol{Q}^\top] \cdot \boldsymbol{Q} + \lambda \tilde{\boldsymbol{W}} \cdot \boldsymbol{Q} \\
&= \mathbb{E}_{\boldsymbol{x},y \sim \mathcal{P}_I} [(f_2(\boldsymbol{Q} \boldsymbol{x}) - onehot(y)) (\boldsymbol{Q} \boldsymbol{x})^\top] \cdot \boldsymbol{Q} + \lambda \tilde{\boldsymbol{W}} \cdot \boldsymbol{Q} \\
&= \mathbb{E}_{\boldsymbol{x},y \sim \tilde{\mathcal{P}}_I} [(f_2(\boldsymbol{x}) - onehot(y)) \boldsymbol{x}^\top] \cdot \boldsymbol{Q} + \lambda \tilde{\boldsymbol{W}} \cdot \boldsymbol{Q} \\
&= \frac{\partial L_2}{\partial \tilde{\boldsymbol{W}}} \cdot \boldsymbol{Q}.
\label{eq:proof_lemma1_grad}
\end{aligned} \end{equation}

After a single optimization step, the gradient decent process of the two classifiers are
\begin{equation} \begin{aligned}
\boldsymbol{W}_{n+1} = \boldsymbol{W}_{n} - \eta \frac{\partial L_1}{\partial \boldsymbol{W}_n}, \\
\tilde{\boldsymbol{W}}_{n+1} = \tilde{\boldsymbol{W}}_{n} - \eta \frac{\partial L_2}{\partial \tilde{\boldsymbol{W}}_n},
\end{aligned} \end{equation}
where $\eta$ is the learning rate.

We can then use \cref{eq:proof_lemma1_grad} to obtain
\begin{equation} \begin{aligned}
\boldsymbol{W}_{n+1} &= \boldsymbol{W}_{n} - \eta \frac{\partial L_1}{\partial \boldsymbol{W}_n}, \\
&= \tilde{\boldsymbol{W}}_n \cdot \boldsymbol{Q} - \eta \frac{\partial L_2}{\partial \tilde{\boldsymbol{W}}_n} \cdot \boldsymbol{Q}, \\
&= (\tilde{\boldsymbol{W}}_n - \eta \frac{\partial L_2}{\partial \tilde{\boldsymbol{W}}_n}) \cdot \boldsymbol{Q}, \\
&= \tilde{\boldsymbol{W}}_{n+1} \cdot \boldsymbol{Q},
\end{aligned} \end{equation}

We have completed this proof.

\subsection{Proof of Proposition 1}
With Lemma 1 established, we can construct an orthogonal matrix $\boldsymbol{Q}$ to obtain the orthogonally transformed distribution $\tilde{\mathcal{P}}_I$, ensuring the Semantic Difference Space and the Covariate Space are decoupled.
The construction process of the orthogonal matrix $\boldsymbol{Q}$ is as follows.

We first apply the Gram-Schmidt orthogonalization to obtain a set of orthonormal basis vectors for the Semantic Difference Space $\mathcal{S}$
\begin{equation} \begin{aligned}
\{\boldsymbol{q}_1,...,\boldsymbol{q}_r\} = GS(\{\boldsymbol{\mu}_1-\boldsymbol{\mu}_2,...,\boldsymbol{\mu}_k-\boldsymbol{\mu}_{k-1}\}),
\end{aligned} \end{equation}
where $GS(\cdot)$ represents the Gram-Schmidt orthogonalization. Note that the input vector set $\{\boldsymbol{\mu}_1-\boldsymbol{\mu}_2,...,\boldsymbol{\mu}_k-\boldsymbol{\mu}_{k-1}\}$ may not be linearly independent, so during the Gram-Schmidt orthogonalization process, zero vectors may appear. We skip these vectors and only retain the non-zero vectors as the basis, denoted as $\{\boldsymbol{q}_1,...,\boldsymbol{q}_r\}$, $r \leq k-1$. Therefore, the number $r$ of basis vectors corresponds to the rank of the Semantic Difference Space.

We then extend this orthonormal basis to $d$ dimensions, yielding a set of orthonormal basis vectors for the input space $\mathcal{X}$.
\begin{equation} \begin{aligned}
\{\boldsymbol{q}_1,...,\boldsymbol{q}_r, \boldsymbol{q}_{r+1}, ..., \boldsymbol{q}_d\} &= GS(\{\boldsymbol{q}_1,...,\boldsymbol{q}_r, \boldsymbol{e}_1,..., \boldsymbol{e}_d\}), \\
\boldsymbol{e}_i &= [0,...,0,1,0,...,0]^\top,
\end{aligned} \end{equation}
where $\boldsymbol{e}_i$ is a standard basis vector with the $i^{th}$ component is 1 and all other components are 0. 
We concatenate the Semantic Difference Space basis vectors $\{\boldsymbol{q}_1,...,\boldsymbol{q}_r\}$ with the standard basis vectors $\{\boldsymbol{e}_1,...,\boldsymbol{e}_d\}$ before the Gram-Schmidt orthogonalization, ensuring the completeness of the final d-dimensional basis.
Since the input space $\mathcal{X}$ is decomposed into the direct sum of subspaces $\mathcal{S}$ and $\mathcal{C}$, it is easy to see that $\{\boldsymbol{q}_{r+1},...,\boldsymbol{q}_d\}$ forms a set of orthonormal basis vectors for the Covariate Space $\mathcal{C}$.

At this point, we obtain the orthonormal bases for both the Semantic Difference Space and the Covariate Space
\begin{equation} \begin{aligned}
span(\{\boldsymbol{q}_1,...,\boldsymbol{q}_r\}) = &span(\{\boldsymbol{\mu}_1-\boldsymbol{\mu}_2,...,\boldsymbol{\mu}_k-\boldsymbol{\mu}_{k-1}\}) = \mathcal{S}, \\
&span(\{\boldsymbol{q}_{r+1}, ..., \boldsymbol{q}_d\}) = \mathcal{C}.
\end{aligned} \end{equation}

Thus, each class center can be expressed in terms of these basis vectors
\begin{equation} \begin{aligned}
\boldsymbol{\mu}_i &= \boldsymbol{s}_i + \boldsymbol{c}_{const}, \boldsymbol{s}_i \in \mathcal{S}, \boldsymbol{c}_{const} \in \mathcal{C}, \\
\boldsymbol{s}_i &= s_{i1}\boldsymbol{q}_1+...+s_{ir}\boldsymbol{q}_r,\\
\boldsymbol{c}_{const} &= c_{r+1}\boldsymbol{q}_{r+1}+...+c_{d}\boldsymbol{q}_d.
\end{aligned} \end{equation}

Based on the basis vectors, we can construct an orthogonal matrix $\boldsymbol{Q}$ to decompose the Semantic Difference Space and the Covariate Space
\begin{equation} \begin{aligned}
\boldsymbol{Q} = [\boldsymbol{q}_1,...,\boldsymbol{q}_r, \boldsymbol{q}_{r+1}, ..., \boldsymbol{q}_d]^\top.
\end{aligned} \end{equation}

At this point, we complete the construction of the orthogonal matrix $\boldsymbol{Q}$, and we obtain the new input distribution $(\boldsymbol{x},y) \sim \tilde{\mathcal{P}}_I$.
After the orthogonal transformation, the ID class centers become
\begin{equation} \begin{aligned}
\tilde{\boldsymbol{\mu}}_i = Q \cdot \boldsymbol{\mu}_i = [s_{i1},...,s_{ir},c_{r+1},...,c_d]^\top, i \in \{1,...,k\}. \\
\end{aligned} \end{equation}

Since a Gaussian distribution remains Gaussian after an orthogonal transformation, the input distribution of each ID class can now be expressed as follows
\begin{equation} \begin{aligned}
\mathcal{N}(\tilde{\boldsymbol{\mu}}_i,\boldsymbol{I}), i \in \{1,...,k\}. \\
\end{aligned} \end{equation}

As a result, after applying Lemma 1, the proof of Proposition 1 reduces to studying the training process of a new linear classifier with a simplified weight matrix $\tilde{\boldsymbol{W}} = \boldsymbol{W} \cdot \boldsymbol{Q}^\top$ within the orthogonally transformed distribution $\tilde{\mathcal{P}}_I$, where the ID distribution follows 
\begin{equation} \begin{aligned}
\tilde{\mathcal{P}}_I|y_i &= (\mathcal{N}(\tilde{\boldsymbol{\mu}}_i,\boldsymbol{I}), y_i), i \in \{1,...,k\}, \\
\tilde{\boldsymbol{\mu}}_i &= [s_{i1},...,s_{ir},c_{r+1},...,c_d]^\top.
\end{aligned} \end{equation}
The last $d - r$ dimensions of the class centers are constants and do not vary with the input label. 

The output of the new linear classifier is now as follows
\begin{equation} \begin{aligned}
f(\boldsymbol{x}) &= softmax(\tilde{\boldsymbol{W}}\boldsymbol{x}) = [\tilde{p}_1(\boldsymbol{x}),...,\tilde{p}_k(\boldsymbol{x})]^\top, \\
\tilde{p}_i(\boldsymbol{x}) &= \frac{\exp{(\tilde{\boldsymbol{W}}_{i,:} \boldsymbol{x})}}{\sum_{j=1}^k \exp{(\tilde{\boldsymbol{W}}_{j,:} \boldsymbol{x})}}, i \in \{1,...,k\}.
\end{aligned} \end{equation}

Considering the training process, the total loss is set as
\begin{equation} \begin{aligned}
Loss = - \mathbb{E}_{\boldsymbol{x},y}[\sum_{i=1}^k \mathbb{I}\{y=i\} log(\tilde{p}_i(\boldsymbol{x}))] + \frac{\lambda}{2} \Vert \tilde{\boldsymbol{W}} \Vert_2^2,
\end{aligned} \end{equation}
where the first term is the cross-entropy loss, and the second term is the L2 regularization loss. $\mathbb{I}\{y=i\}$ represents the indicator, which equals 1 when $y=i$, and 0 otherwise.
The partial derivative of the loss function with respect to the classifier's weight of class $i$ on a specific input dimension $j$ is then computed as follows
\begin{equation} \begin{aligned}
\frac{\partial Loss}{\partial \tilde{\boldsymbol{W}}_{ij}} = \mathbb{E}_{\boldsymbol{x},y} [(\tilde{p}_i(\boldsymbol{x}) - \mathbb{I}\{y=i\}) \boldsymbol{x}_j] + \lambda \tilde{\boldsymbol{W}}_{ij}.
\label{eq:proof_pro2_loss_grad}
\end{aligned} \end{equation}

Now we examine the last $d-r$ columns of the simplified matrix $\tilde{\boldsymbol{W}}$. Since when $j>r$, the value of $\boldsymbol{x}_j$ is independent with the input label $y$. We can derive the first term of \cref{eq:proof_pro2_loss_grad} as follows
\begin{equation} \begin{aligned}
\mathbb{E}_{\boldsymbol{x},y} [(\tilde{p}_i(\boldsymbol{x}) - \mathbb{I}\{y=i\}) \boldsymbol{x}_j] 
&= \mathbb{E}_{\boldsymbol{x},y} [\tilde{p}_i(\boldsymbol{x})\boldsymbol{x}_j] - \mathbb{E}_{\boldsymbol{x},y} [\mathbb{I}\{y=i\}\boldsymbol{x}_j] \\
&= \mathbb{E}_{\boldsymbol{x},y} [\tilde{p}_i(\boldsymbol{x})\boldsymbol{x}_j] - \mathbb{E}_{y} [\mathbb{I}\{y=i\}]\mathbb{E}_{\boldsymbol{x}}[\boldsymbol{x}_j] \\
&= \mathbb{E}_{\boldsymbol{x},y} [\tilde{p}_i(\boldsymbol{x})\boldsymbol{x}_j] - \frac{1}{k} \cdot c_j \\
&= \mathbb{E}_{\boldsymbol{x},y} [\tilde{p}_i(\boldsymbol{x})]\mathbb{E}_{\boldsymbol{x},y}[\boldsymbol{x}_j] \\
&\quad + Cov(\tilde{p}_i(\boldsymbol{x}),\boldsymbol{x}_j) - \frac{c_j}{k}
\end{aligned} \end{equation}

Applying Assumption 1, we obtain
\begin{equation} \begin{aligned}
\mathbb{E}_{\boldsymbol{x},y} [(\tilde{p}_i(\boldsymbol{x}) - \mathbb{I}\{y=i\}) \boldsymbol{x}_j] 
&= \mathbb{E}_{\boldsymbol{x},y} [\tilde{p}_i(\boldsymbol{x})]\mathbb{E}_{\boldsymbol{x},y}[\boldsymbol{x}_j] \\
&\quad + Cov(\tilde{p}_i(\boldsymbol{x}),\boldsymbol{x}_j) - \frac{c_j}{k} \\
&= \frac{1}{k} \cdot c_j + Cov(\tilde{p}_i(\boldsymbol{x}),\boldsymbol{x}_j) - \frac{c_j}{k} \\
&= Cov(\tilde{p}_i(\boldsymbol{x}),\boldsymbol{x}_j)
\end{aligned} \end{equation}

Substituting this result back into \cref{eq:proof_pro2_loss_grad}, the expression for the partial derivative becomes
\begin{equation} \begin{aligned}
\frac{\partial Loss}{\partial \tilde{\boldsymbol{W}}_{ij}} = Cov(\tilde{p}_i(\boldsymbol{x}),\boldsymbol{x}_j) + \lambda \tilde{\boldsymbol{W}}_{ij}.
\end{aligned} \end{equation}

We multiply the partial derivative by the weight $\tilde{\boldsymbol{W}}_{ij}$ itself to analyze the direction of the weight update
\begin{equation} \begin{aligned}
\tilde{\boldsymbol{W}}_{ij} \cdot \frac{\partial Loss}{\partial \tilde{\boldsymbol{W}}_{ij}} = \tilde{\boldsymbol{W}}_{ij} \cdot Cov(\tilde{p}_i(\boldsymbol{x}),\boldsymbol{x}_j) + \lambda \tilde{\boldsymbol{W}}_{ij}^2.
\end{aligned} \end{equation}

After applying Assumption 2,
\begin{equation} \begin{aligned}
\tilde{\boldsymbol{W}}_{ij} \cdot \frac{\partial Loss}{\partial \tilde{\boldsymbol{W}}_{ij}} \geq \lambda \tilde{\boldsymbol{W}}_{ij}^2 \geq 0.
\label{eq:proof_pro2_over_zero}
\end{aligned} \end{equation}

The derivation shows that for a given weight $\tilde{\boldsymbol{W}}_{ij}$ of class $i$ on the $j^{th}$ input dimension, $j>r$, the partial derivative of the loss function with respect to that weight has the same sign as the weight itself. This means that the absolute value of the weight will continuously decrease throughout the training process due to gradient descent optimization. 

Next, we will further prove that $\tilde{\boldsymbol{W}}_{ij}$ converges to 0 when $j>r$. Considering the gradient descent process
\begin{equation} \begin{aligned}
\tilde{\boldsymbol{W}}_{n+1,ij} = \tilde{\boldsymbol{W}}_{n,ij} - \eta \frac{\partial Loss}{\partial \tilde{\boldsymbol{W}}_{n,ij}},
\label{eq:proof_pro2_grad_descent}
\end{aligned} \end{equation}
where $n$ denotes the training steps, $\tilde{\boldsymbol{W}}_{n,ij}$ is the weight in training step $n$, and $\eta$ refers to the learning rate. In the training setup, the learning rate $\eta$ is set to a small value to ensure the convergence of the training process. Therefore, we assume in the training process,
\begin{equation} \begin{aligned}
|\tilde{\boldsymbol{W}}_{n,ij}| \geq |\eta \frac{\partial Loss}{\partial \tilde{\boldsymbol{W}}_{n,ij}}|.
\end{aligned} \end{equation}
Combined with \cref{eq:proof_pro2_over_zero}, we obtain
\begin{equation} \begin{aligned}
\tilde{\boldsymbol{W}}_{n,ij} \cdot \eta \frac{\partial Loss}{\partial \tilde{\boldsymbol{W}}_{n,ij}} \geq (\eta \frac{\partial Loss}{\partial \tilde{\boldsymbol{W}}_{n,ij}})^2.
\label{eq:proof_pro2_mid}
\end{aligned} \end{equation}

Squaring both sides of \cref{eq:proof_pro2_grad_descent} and substituting \cref{eq:proof_pro2_over_zero} and \cref{eq:proof_pro2_mid},
\begin{equation} \begin{aligned}
\tilde{\boldsymbol{W}}_{n+1,ij}^2 &= \tilde{\boldsymbol{W}}_{n,ij}^2 + (\eta \frac{\partial Loss}{\partial \tilde{\boldsymbol{W}}_{n,ij}})^2 - 2\tilde{\boldsymbol{W}}_{n,ij} \cdot \eta \frac{\partial Loss}{\partial \tilde{\boldsymbol{W}}_{n,ij}} \\
&= \tilde{\boldsymbol{W}}_{n,ij}^2 + ((\eta \frac{\partial Loss}{\partial \tilde{\boldsymbol{W}}_{n,ij}})^2 \\
&\quad - \tilde{\boldsymbol{W}}_{n,ij} \cdot \eta \frac{\partial Loss}{\partial \tilde{\boldsymbol{W}}_{n,ij}}) - \tilde{\boldsymbol{W}}_{n,ij} \cdot \eta \frac{\partial Loss}{\partial \tilde{\boldsymbol{W}}_{n,ij}} \\
&\leq \tilde{\boldsymbol{W}}_{n,ij}^2 - \tilde{\boldsymbol{W}}_{n,ij} \cdot \eta \frac{\partial Loss}{\partial \tilde{\boldsymbol{W}}_{n,ij}} \\
&= \tilde{\boldsymbol{W}}_{n,ij}^2 - \eta \cdot (\tilde{\boldsymbol{W}}_{n,ij} \cdot \frac{\partial Loss}{\partial \tilde{\boldsymbol{W}}_{n,ij}}) \\
&\leq \tilde{\boldsymbol{W}}_{n,ij}^2 - \eta \cdot \lambda \tilde{\boldsymbol{W}}_{ij}^2 \\
&= (1-\eta \lambda)\tilde{\boldsymbol{W}}_{n,ij}^2
\end{aligned} \end{equation}

Thus, the final convergence result is
\begin{equation} \begin{aligned}
|\tilde{\boldsymbol{W}}_{n+1,ij}| \leq (1-\eta \lambda)^{\frac{1}{2}} |\tilde{\boldsymbol{W}}_{n,ij}|
\leq (1-\eta \lambda)^{\frac{n}{2}} |\tilde{\boldsymbol{W}}_{1,ij}|
\xrightarrow{n \to \infty} 0.
\end{aligned} \end{equation}
The equation holds because $\eta$ and $\lambda$ are set to small positive values, which ensures $0<(1-\eta \lambda)<1$.

The result implies that once the training has converged, $\tilde{\boldsymbol{W}}_{ij}$ converges to 0 when $j>r$, which means
\begin{equation} \begin{aligned}
\tilde{\boldsymbol{W}}_{:,r+1} = ... = \tilde{\boldsymbol{W}}_{:,d} = \boldsymbol{0}.
\end{aligned} \end{equation}

We have completed this proof.

\subsection{Proof of Theorem 1}
Under the linear assumption, we can obtain
\begin{equation} \begin{aligned}
proj_\mathcal{S}(\boldsymbol{\mu}_a-\boldsymbol{\mu}_b) = proj_\mathcal{S}(\boldsymbol{\mu}_a) - proj_\mathcal{S}(\boldsymbol{\mu}_b) = \mathbf{0}
\end{aligned} \end{equation}

Since the centers of the two classes $\mathcal{N}(\boldsymbol{\mu}_a,\boldsymbol{I})$ and $\mathcal{N}(\boldsymbol{\mu}_b,\boldsymbol{I})$ are identical upon projection into the Semantic Difference Space $\mathcal{S}$, we assume the decomposition of the two class centers is as follows
\begin{equation} \begin{aligned}
\boldsymbol{\mu}_a &= \boldsymbol{s} + \boldsymbol{c}_a, \\
\boldsymbol{\mu}_b &= \boldsymbol{s} + \boldsymbol{c}_b, \\
\boldsymbol{s} \in \mathcal{S}, &\quad \boldsymbol{c}_a,\boldsymbol{c}_b \in \mathcal{C}. 
\end{aligned} \end{equation}

Using the same method from the proof of Proposition 1 to construct the orthogonal basis vectors and the orthogonal matrix, we can express the two class centers in terms of the basis
\begin{equation} \begin{aligned}
\boldsymbol{\mu}_a &= s_1 \boldsymbol{q}_1+...+s_r \boldsymbol{q}_r+c_{a,r+1} \boldsymbol{q}_{r+1}+...+c_{a,d} \boldsymbol{q}_d, \\
\boldsymbol{\mu}_b &= s_1 \boldsymbol{q}_1+...+s_r \boldsymbol{q}_r+c_{b,r+1} \boldsymbol{q}_{r+1}+...+c_{b,d} \boldsymbol{q}_d. \\
\end{aligned} \end{equation}

According to the linear transformation of Gaussians, we can derive the output distribution $\mathcal{P}_{a}$ of the input distribution $\mathcal{N}(\boldsymbol{\mu}_a,\boldsymbol{I})$ after being transformed by the weight matrix $\boldsymbol{W}$
\begin{equation} \begin{aligned}
\boldsymbol{o}_a &= \boldsymbol{W}\boldsymbol{x}_a, \\
\boldsymbol{x} &\sim \mathcal{N}(\boldsymbol{\mu}_a,\boldsymbol{I}), \\
\boldsymbol{o}_a \sim \mathcal{P}_{a} &= \mathcal{N}(\boldsymbol{W}\boldsymbol{\mu}_a,\boldsymbol{W}\boldsymbol{I}\boldsymbol{W}^\top).
\end{aligned} \end{equation}

Based on Proposition 1 that $\boldsymbol{W} = \tilde{\boldsymbol{W}} \cdot \boldsymbol{Q}$, we can conclude
\begin{equation} \begin{aligned}
\boldsymbol{W}\boldsymbol{\mu}_a &= \tilde{\boldsymbol{W}} \boldsymbol{Q} \boldsymbol{\mu}_a \\
&= \tilde{\boldsymbol{W}} [\boldsymbol{q}_1,...,\boldsymbol{q}_r, \boldsymbol{q}_{r+1}, ..., \boldsymbol{q}_d]^\top \\
&\quad (s_1 \boldsymbol{q}_1+...+s_r \boldsymbol{q}_r+c_{a,r+1} \boldsymbol{q}_{r+1}+...+c_{a,d} \boldsymbol{q}_d) \\
&= \tilde{\boldsymbol{W}} [s_1,...,s_r,c_{a,r+1},...,c_{a,d}]^\top \\
&= [\tilde{\boldsymbol{W}}_{:,1},...,\tilde{\boldsymbol{W}}_{:,r},0,...,0] [s_1,...,s_r,c_{a,r+1},...,c_{a,d}]^\top \\
&= s_1\tilde{\boldsymbol{W}}_{:,1}+...+s_r\tilde{\boldsymbol{W}}_{:,r}. \\
\end{aligned} \end{equation}

This implies the output distribution of the input distribution $\mathcal{N}(\boldsymbol{\mu}_a,\boldsymbol{I})$ is
\begin{equation} \begin{aligned}
\boldsymbol{o}_a \sim \mathcal{P}_{a} = \mathcal{N}(s_1\tilde{\boldsymbol{W}}_{:,1}+...+s_r\tilde{\boldsymbol{W}}_{:,r},\boldsymbol{W}\boldsymbol{I}\boldsymbol{W}^\top). \\
\end{aligned} \end{equation}
It can be observed that the output distribution is independent of the Covariate Space component $\boldsymbol{c}_a$ of the input's class center $\boldsymbol{\mu}_a$.

Using the same method, we can obtain the output distribution of the other input distribution $\mathcal{N}(\boldsymbol{\mu}_b,\boldsymbol{I})$
\begin{equation} \begin{aligned}
\boldsymbol{o}_b \sim \mathcal{P}_{b} = \mathcal{N}(s_1\tilde{\boldsymbol{W}}_{:,1}+...+s_r\tilde{\boldsymbol{W}}_{:,r},\boldsymbol{W}\boldsymbol{I}\boldsymbol{W}^\top),
\end{aligned} \end{equation}
which is the same as the output of $\mathcal{N}(\boldsymbol{\mu}_a,\boldsymbol{I})$. It implies
\begin{equation} \begin{aligned}
KL(\mathcal{P}_{a}||\mathcal{P}_{b})=0.
\end{aligned} \end{equation}

Since the classifier output just includes an additional softmax layer after applying the weight matrix $\boldsymbol{W}$
\begin{equation} \begin{aligned}
f(\boldsymbol{x}) = softmax(\boldsymbol{W}\boldsymbol{x}) = softmax(\boldsymbol{o}).
\end{aligned} \end{equation}

Thus, the KL divergence between the classifier's output of the two distributions is
\begin{equation} \begin{aligned}
&KL(f(\mathcal{N}(\boldsymbol{\mu}_a,\boldsymbol{I})) || f(\mathcal{N}(\boldsymbol{\mu}_b,\boldsymbol{I}))) \\
= &KL(softmax(\mathcal{P}_{a}) || softmax(\mathcal{P}_{b})) \\
= &0
\end{aligned} \end{equation}

We have completed this proof.

\subsection{Necessary Definition and Lemma for Theorem 2}

For the convenience of the proof, we have made some formal definitions of the properties of the loss function as follows:

\textbf{Definition 4.} \textit{The loss function $L$ satisfies the following properties: \\
1. $L(f,g) \geq 0$, and $L(f,g) = 0$ iff $f(x_s, x_c) = g(x_s)$; \\
2. $L(f,g)$ increases monotonically with $|f - g|$, and for any input region $\Delta \mathcal{X}$, there exists a constant $k_{\Delta \mathcal{X}}$ such that $E_{\Delta \mathcal{X}}L(f,g) \geq k_{\Delta \mathcal{X}}E_{\Delta \mathcal{X}}|f - g|$.}
\label{def:loss_function}

The first property ensures that the loss function can converge to a minimum value, and that it attains the minimum only when the model is optimized to match the target function. We set the minimum value of the loss to 0 for convenient. The second property ensures that the loss effectively measures the difference between the model function and the target function. The coefficient $k_{\Delta \mathcal{X}}$ is imposed to ensure that the loss function is not an insignificant higher-order small quantity during training. We find that common loss functions, such as cross-entropy (CE), satisfy the above properties.

\textbf{Lemma 2.} \textit{If for an input region $\Delta \mathcal{X} = \{x|a \leq x \leq b\}$ where $\frac{\partial f}{\partial x}$ is continuous and $|\frac{\partial f}{\partial x}| > \epsilon$, then we have $\int_{\Delta \mathcal{X}} |f(x)| dx > \frac{\epsilon}{4}(b-a)^2$.}

\textbf{Proof of Lemma 2.}
Since $\frac{\partial f}{\partial x}$ is continuous and $|\frac{\partial f}{\partial x}| > \epsilon$, we have $\frac{\partial f}{\partial x} > \epsilon$ or $\frac{\partial f}{\partial x} < -\epsilon$. By symmetry, we can assume $\frac{\partial f}{\partial x} > \epsilon$ in the proof.

We consider three cases:

1. $f(x)>0$ on $\Delta \mathcal{X}$. We have:
\begin{equation} \begin{aligned}
\int_{\Delta \mathcal{X}} |f(x)| dx &= \int_{\Delta \mathcal{X}} f(x) dx \\
                                    &> \int_{\Delta \mathcal{X}} [f(a) + \epsilon(x-a)] dx \\
                                    &= f(a)(b-a) + \frac{\epsilon}{2}(b-a)^2 \\
                                    &> \frac{\epsilon}{2}(b-a)^2
\end{aligned} \end{equation}

2. $f(x)<0$ on $\Delta \mathcal{X}$. We have:
\begin{equation} \begin{aligned}
\int_{\Delta \mathcal{X}} |f(x)| dx &= \int_{\Delta \mathcal{X}} -f(x) dx \\
                                    &> \int_{\Delta \mathcal{X}} [-f(b) + \epsilon(b-x)] dx \\
                                    &= -f(b)(b-a) + \frac{\epsilon}{2}(b-a)^2 \\
                                    &> \frac{\epsilon}{2}(b-a)^2
\end{aligned} \end{equation}

3. There exist $c \in \Delta \mathcal{X}$ satisfies $f(c)=0$. We have:
\begin{equation} \begin{aligned}
\int_{\Delta \mathcal{X}} |f(x)| dx &= \int_{a}^c -f(x) dx + \int_{c}^b f(x) dx \\
                                    &> \int_{a}^c \epsilon(c-x) dx + \int_{c}^b \epsilon(x-c) dx \\
                                    &= \frac{\epsilon}{2}(c-a)^2 + \frac{\epsilon}{2}(b-c)^2 \\
                                    &\geq \frac{\epsilon}{4}(b-a)^2
\end{aligned} \end{equation}

By considering the three cases together, we obtain:
\begin{equation} \begin{aligned}
\int_{\Delta \mathcal{X}} |f(x)| dx > \frac{\epsilon}{4}(b-a)^2
\end{aligned} \end{equation}

We have completed the proof.

\subsection{Proof of Theorem 2}

To prove by contradiction, we need to demonstrate that the condition does not hold when:
\begin{equation} \begin{aligned}
|\frac{\partial f}{\partial c_j}| > \frac{4\epsilon_{loss}}{P_{\Delta \mathcal{X}} k_{\Delta \mathcal{X}} p_m (b_{c,j}-a_{c,j})^2}
\label{eq:nonlinear_contradiction}
\end{aligned} \end{equation}

Based on the first property ($L(f,g) \geq 0$) in Definition 4, we get:
\begin{equation} \begin{aligned}
E_{\mathcal{X}}L(f,g) &= \int_{\mathcal{X}}L(f,g)p(x)dx \\
                        &\geq \int_{\Delta \mathcal{X}}L(f,g)p(x)dx \\
                        &= P_{\Delta \mathcal{X}}E_{\Delta \mathcal{X}}L(f,g)
\label{eq:nonlinear_1}
\end{aligned} \end{equation}

We can further use the second property to obtain:
\begin{equation} \begin{aligned}
E_{\Delta \mathcal{X}}L(f,g)  \geq k_{\Delta \mathcal{X}}E_{\Delta \mathcal{X}}|f - g|
\label{eq:nonlinear_2}
\end{aligned} \end{equation}

Since the features are independent of each other, the expectation can be written in the following form:
\begin{equation} \begin{aligned}
E_{\Delta \mathcal{X}}|f - g| =  &E_{s_1 \in [a_{s,1}, b_{s,1}]} ... E_{s_r \in [a_{s,r}, b_{s,r}]}[\\
&E_{c_1 \in [a_{c,1}, b_{c,1}]} ... E_{c_{d-r} \in [a_{c,d-r}, b_{c,d-r}]}|f - g|]
\label{eq:nonlinear_3}
\end{aligned} \end{equation}

We can first calculate the expectation on $c_j$ and apply the condition of probability density $p(c_j) \geq p_m$:
\begin{equation} \begin{aligned}
E_{c_j \in [a_{c,j}, b_{c,j}]}|f - g| &= \int_{a_{c,j}}^{b_{c,j}} |f - g| p(c_j)dc_j \\
                        & \geq p_m\int_{a_{c,j}}^{b_{c,j}} |f - g| dc_j 
\label{eq:nonlinear_4}
\end{aligned} \end{equation}

Since $\frac{\partial f}{\partial c_j}$ is constant throughout \(\Delta\mathcal X\), we can apply Lemma 2 using the assumption in \cref{eq:nonlinear_contradiction} to obtain:
\begin{equation} \begin{aligned}
\int_{a_{c,j}}^{b_{c,j}} |f - g| dc_j &> \frac{1}{4} \frac{4\epsilon_{loss}}{P_{\Delta \mathcal{X}} k_{\Delta \mathcal{X}} p_m (b_{c,j}-a_{c,j})^2} (b_{c,j}-a_{c,j})^2\\
&= \frac{\epsilon_{loss}}{P_{\Delta \mathcal{X}} k_{\Delta \mathcal{X}} p_m }
\label{eq:nonlinear_5}
\end{aligned} \end{equation}

At this point, we obtain a constant lower bound for the expectation on $c_j$ using \cref{eq:nonlinear_4} and \cref{eq:nonlinear_5}:
\begin{equation} \begin{aligned}
E_{c_j \in [a_{c,j}, b_{c,j}]}|f - g| > \frac{\epsilon_{loss}}{P_{\Delta \mathcal{X}} k_{\Delta \mathcal{X}}}
\end{aligned} \end{equation}

Since the expectation is independent of $(x_s,x_c)$, we can substitute it into \cref{eq:nonlinear_3} to eliminate the calculation of other expectations, thereby obtaining the lower bound for the overall expectation:
\begin{equation} \begin{aligned}
E_{\Delta \mathcal{X}}|f - g| > \frac{\epsilon_{loss}}{P_{\Delta \mathcal{X}} k_{\Delta \mathcal{X}}}
\label{eq:nonlinear_6}
\end{aligned} \end{equation}

By combining \cref{eq:nonlinear_1}, \cref{eq:nonlinear_2}, and \cref{eq:nonlinear_6}, we obtain:
\begin{equation} \begin{aligned}
E_{\mathcal{X}}L(f,g) > \epsilon_{loss}
\end{aligned} \end{equation}

However, the condition states that:
\begin{equation} \begin{aligned}
E_{\mathcal{X}}L(f,g) < \epsilon_{loss}
\end{aligned} \end{equation}

This implies that the assumption in \cref{eq:nonlinear_contradiction} does not hold. Therefore, by contradiction, we have completed the proof.

\newpage

\section{Experiments on ViT-Based Architectures}
\label{app:vit}

The theoretical framework developed in this paper is grounded in the analysis of how training objectives shape the model's perceptual space, which applies to any architecture that converges under supervised classification loss. To verify that the OOD separability assessment generalizes beyond ResNet-50, we replicate the existing-benchmark evaluation using ViT-B/16 as the backbone.

\subsection{Experimental Details}

We use the official ImageNet-1K pre-trained ViT-B/16 checkpoint and evaluate the same eight OOD datasets with the same 10 post-hoc detection methods. OSP scores are computed in the same manner as described in the main paper within the feature space of pre-trained ViT-B/16.

\subsection{Results on OOD Separability}

\begin{table*}[t]
\centering
\caption{AUROC (\%) of OOD detection methods and OOD Separability on existing benchmark datasets with ViT-B/16 backbone (ImageNet-1K as ID). The experimental protocol mirrors \cref{tab:exp_IN1K_full} in the main paper.}\label{tab:app_vit_in1k}
\resizebox{1.0\textwidth}{!}{
\renewcommand{\arraystretch}{1.2}
\begin{tabular}{c|c|c|c|c|c|c|c|c}
\toprule
         & \textbf{SSB-hard} & \textbf{NINCO} & \textbf{Textures} & \textbf{iNaturalist} & \textbf{OpenImage-O} & \textbf{ImageNet-V2} & \textbf{ImageNet-C} & \textbf{ImageNet-R} \\
\midrule
MSP~\cite{MSP_2017}                 & 68.9    & 78.7  & 85.3     & 88.0        & 84.6       & 57.7       & 74.8      & 77.9      \\
ODIN~\cite{ODIN_2018}               & 57.7    & 56.9  & 70.1     & 70.5        & 62.0       & 50.7       & 66.7      & 57.7      \\
EBO~\cite{EBO_2020}                 & 59.0    & 66.8  & 81.1     & 79.0        & 76.3       & 54.5       & 73.9      & 73.0      \\
GradNorm~\cite{GradNorm_2021}       & 43.2    & 35.9  & 45.2     & 42.0        & 38.1       & 45.9       & 55.6      & 38.7      \\
RMDS~\cite{RMDS_2021}               & 72.7    & 87.5  & 89.4     & 96.1        & 92.2       & 59.1       & 75.9      & 84.9      \\
KNN~\cite{KNN_2022}                 & 67.4    & 83.1  & 90.2     & 90.0        & 88.6       & 58.7       & 77.2      & 86.3      \\
DICE~\cite{DICE_2022}               & 59.9    & 72.0  & 82.1     & 82.7        & 82.0       & 56.0       & 73.8      & 81.3      \\
ASH~\cite{ASH_2023}                 & 54.1    & 53.4  & 47.0     & 50.6        & 56.0       & 51.4       & 49.1      & 51.2      \\
Relation~\cite{Relation_2023}       & 66.3    & 82.2  & 90.6     & 91.4        & 89.4       & 58.0       & 76.1      & 83.2      \\
SCALE~\cite{SCALE_2024}             & 55.2    & 59.1  & 76.8     & 69.1        & 69.5       & 52.5       & 72.1      & 69.3      \\
\midrule
$S(\mathcal{P}_I, \mathcal{P}_O)$ & 96.3    & 96.6  & 98.4     & 99.7        & 97.6       & 62.4       & 96.5      & 97.5 \\
\bottomrule
\end{tabular}
}
\end{table*}

The results in \cref{tab:app_vit_in1k} reveal a consistent pattern across many experimental settings. Specifically, the OSP trained on the ViT feature space aligns closely with the ResNet-50 results in \cref{tab:exp_IN1K_full}, yielding high separability on all evaluated datasets except ImageNet-V2. Similarly, logit- and feature-space-based OOD detection methods (e.g., MSP and KNN) perform consistently under both ViT and ResNet-50 settings. Such cross-architecture consistency demonstrates that different models trained under identical ID settings learn feature spaces with similar properties. This strongly supports the broad generalizability of our theoretical analysis, and confirms that our OSP method is highly adaptable to feature spaces extracted by diverse network architectures.

In addition, we also find that gradient-based methods like GradNorm experience a noticeable performance drop under the ViT setting. This decline is likely because their gradient-scoring mechanisms are not specifically adapted to the unique gradient behaviors of ViT models.


\section{Experiments on Training Size of OSP}
\label{app:osp_sensitivity}

The OSP is trained on a subset of ID and OOD data for each benchmark setting. To verify that the OSP scores are robust to the choice of training size, we conduct an ablation study varying the proportion of OOD data used for probe training.

\subsection{Experimental Setup}

Using the ImageNet-1K benchmark settings from \cref{tab:exp_IN1K_full} of the main paper, we vary the subset size of OOD data used for OSP training across \{10\%, 20\%, 30\%, 40\%, 50\%\} of the available OOD dataset, while keeping all other settings identical. We report the resulting OSP scores for each OOD dataset.

\begin{table*}[t]
\centering
\caption{OSP scores (\%) under different OOD data proportions used for probe training. Results show the robustness of OSP to the choice of training data size.}\label{tab:app_osp_sensitivity}
\resizebox{1.0\textwidth}{!}{
\renewcommand{\arraystretch}{1.2}
\begin{tabular}{c|c|c|c|c|c|c|c|c}
\toprule
         & \textbf{SSB-hard} & \textbf{NINCO} & \textbf{Textures} & \textbf{iNaturalist} & \textbf{OpenImage-O} & \textbf{ImageNet-V2} & \textbf{ImageNet-C} & \textbf{ImageNet-R} \\
\midrule
10\% & 95.4 & 89.9 & 98.4 & 99.7 & 95.3 & 60.5 & 97.5 & 95.4 \\
20\% & 96.0 & 92.2 & 98.8 & 99.7 & 96.3 & 62.7 & 98.0 & 95.8 \\
30\% & 96.5 & 92.7 & 99.0 & 99.7 & 96.6 & 64.0 & 98.3 & 96.1 \\
40\% & 96.6 & 93.5 & 99.1 & 99.8 & 96.9 & 65.6 & 98.6 & 96.3 \\
50\% & 96.8 & 93.6 & 99.3 & 99.8 & 97.1 & 66.5 & 98.7 & 96.4 \\
\bottomrule
\end{tabular}
}
\end{table*}

\subsection{Analysis}

The results in \cref{tab:app_osp_sensitivity} show that OSP scores remain stable across different training sizes. Decreasing the training size to 10\% OOD data results in no substantial variation in the OSP score, indicating that OSP provides a reliable estimate of benchmark separability even with limited training data.


\section{Linear vs.\ Nonlinear Probe Ablation}
\label{app:probe_ablation}

A natural question is whether the linear probe used in OSP may miss separability patterns that a nonlinear probe could capture. To address this, we compare the linear probe with a 2-layer MLP probe (hidden dim = 2048, ReLU activation) on the same ImageNet-1K benchmark settings as in \cref{tab:exp_IN1K_full}.

\begin{table*}[t]
\centering
\caption{OSP scores (\%) under linear probe and 2-layer MLP probe.}\label{tab:app_probe_ablation}
\resizebox{1.0\textwidth}{!}{
\renewcommand{\arraystretch}{1.2}
\begin{tabular}{c|c|c|c|c|c|c|c|c}
\toprule
         & \textbf{SSB-hard} & \textbf{NINCO} & \textbf{Textures} & \textbf{iNaturalist} & \textbf{OpenImage-O} & \textbf{ImageNet-V2} & \textbf{ImageNet-C} & \textbf{ImageNet-R} \\
\midrule
Linear & 96.8 & 93.6 & 99.3 & 99.8 & 97.1 & 66.5 & 98.7 & 96.4 \\
MLP    & 98.6 & 96.2 & 99.6 & 99.9 & 98.4 & 68.9 & 99.4 & 98.1 \\
\bottomrule
\end{tabular}
}
\end{table*}

The results in \cref{tab:app_probe_ablation} show that the MLP probe yields modestly higher OSP scores across all settings, consistent with its greater representational capacity. However, the relative ranking of benchmark difficulty is identical between the two probe types: iNaturalist and Textures remain the easiest settings, ImageNet-V2 remains the most challenging. This confirms that the linear probe captures the essential separability structure, and that the conclusions drawn from OSP scores are not artifacts of the probe choice.


\section{Experiments on Random Class Splits}
\label{app:splits}

The main experiments on DomainNet OOD and ImageNet Dogs use a fixed random split to partition ID and OOD classes. To verify that the reported conclusions are not artifacts of a particular split, we conduct additional experiments with different random splits.

\subsection{DomainNet OOD}

In the main paper, we fix 300 out of 350 object classes as ID and the remaining 50 as OOD. Here we repeat this procedure with 3 different random seeds to obtain 3 independent splits. For each split, we retrain the model and report AUROC results for post-hoc OOD detection methods.

\begin{table*}[t]
\centering
\caption{AUROC (\%) of representative OOD detection methods on DomainNet OOD under 3 random object class splits (Quickdraw as OOD domain). Mean and standard deviation are reported.}\label{tab:app_domainnet_splits}
\resizebox{0.7\textwidth}{!}{
\renewcommand{\arraystretch}{1.2}
\begin{tabular}{c|c|c|c|c}
\toprule
Testing Protocol & \multicolumn{2}{c|}{OOD-object} & \multicolumn{2}{c}{OOD-domain} \\
\midrule
Training Perspective & object & domain & object & domain \\
\midrule
MSP~\cite{MSP_2017}           & 74.5±0.8 & 52.3±1.9 & 80.3±0.5 & 94.0±1.3  \\
ODIN~\cite{ODIN_2018}         & 72.5±0.8 & 51.8±1.1 & 94.0±0.7 & 97.0±0.4  \\
EBO~\cite{EBO_2020}           & 73.7±0.9 & 52.8±1.8 & 85.3±1.6 & 88.0±1.2  \\
GradNorm~\cite{GradNorm_2021} & 62.0±0.8 & 53.2±1.7 & 77.5±3.7 & 96.9±0.1  \\
RMDS~\cite{RMDS_2021}         & 73.0±0.4 & 51.0±0.9 & 74.9±0.9 & 89.0±1.9  \\
KNN~\cite{KNN_2022}           & 65.6±1.6 & 52.0±1.6 & 87.0±1.5 & 51.2±16.6 \\
DICE~\cite{DICE_2022}         & 68.6±1.0 & 52.2±1.8 & 84.6±2.5 & 95.9±2.7  \\
ASH~\cite{ASH_2023}           & 71.2±1.4 & 52.9±1.8 & 82.6±1.3 & 93.8±1.8  \\
Relation~\cite{Relation_2023} & 66.8±1.5 & 49.9±0.8 & 80.0±0.8 & 82.4±9.9  \\
SCALE~\cite{SCALE_2024}       & 72.2±1.3 & 53.0±1.7 & 83.4±1.2 & 96.3±0.4  \\
\bottomrule
\end{tabular}
}
\end{table*}

The results in \cref{tab:app_domainnet_splits} confirm the key conclusions from the main paper: perspective-matched training leads to better OOD detection, and KNN exhibits anomalous behavior under domain-separated training.

\subsection{ImageNet Dogs}

In the main paper, we fix ID/OOD dog classes as well as the ID/OOD object classes using a single random seed. Here we repeat this procedure with 3 different random seeds to obtain 3 independent splits.

\begin{table*}[t]
\centering
\caption{AUROC (\%) on the OOD-breed testing protocol under 3 random splits. Mean and standard deviation are reported.}\label{tab:app_dog_splits}
\resizebox{0.7\textwidth}{!}{
\renewcommand{\arraystretch}{1.2}
\begin{tabular}{c|c|c|c|c}
\toprule
Testing Protocol                & \multicolumn{2}{c|}{OOD-object} & \multicolumn{2}{c}{OOD-breed} \\
\midrule
Training Granularity              & aggregated & separated       & aggregated & separated      \\
\midrule
\multicolumn{5}{c}{Post-hoc Method} \\
\midrule
MSP~\cite{MSP_2017}           & 82.0±1.1 & 94.1±0.8 & 50.1±0.6 & 73.9±0.2 \\
ODIN~\cite{ODIN_2018}         & 81.1±1.0 & 95.9±0.3 & 51.4±0.7 & 75.0±0.5 \\
EBO~\cite{EBO_2020}           & 81.0±0.8 & 97.3±0.3 & 50.6±0.8 & 75.6±0.6 \\
GradNorm~\cite{GradNorm_2021} & 63.6±2.2 & 93.2±1.9 & 51.9±2.3 & 61.3±1.3 \\
RMDS~\cite{RMDS_2021}         & 81.0±1.6 & 94.0±0.2 & 48.2±2.4 & 76.1±1.3 \\
KNN~\cite{KNN_2022}           & 74.9±0.7 & 97.4±0.7 & 49.4±1.2 & 72.6±2.6 \\
DICE~\cite{DICE_2022}         & 77.2±1.3 & 95.7±1.0 & 51.4±1.0 & 70.0±1.0 \\
ASH~\cite{ASH_2023}           & 77.8±1.6 & 96.6±0.9 & 51.5±0.9 & 71.7±0.3 \\
Relation~\cite{Relation_2023} & 79.3±0.5 & 96.4±0.4 & 50.6±0.4 & 71.7±1.9 \\
SCALE~\cite{SCALE_2024}       & 79.5±1.0 & 97.9±0.2 & 51.9±1.1 & 74.8±1.3 \\
\bottomrule
\end{tabular}
}
\end{table*}

The results in \cref{tab:app_dog_splits} confirm that the granularity effect is robust to the choice of OOD breed split: under \textit{breed-aggregated} training, all methods consistently achieve near-chance AUROC (~50\%) to detect the OOD breeds regardless of which 26 breeds are held out as OOD; under \textit{breed-separated} training, all methods consistently achieve substantially higher AUROC, confirming that the observed effect is a property of the training granularity rather than a particular breed selection.

\revise{
\section{Additional Repeated Experiments on ImageNet Dogs}

To further assess the robustness of the ImageNet Dogs results, we repeat the experiment in \cref{tab:exp_IN_dog} with a total of nine independent training seeds, keeping the ID/OOD class split and all other settings unchanged. \cref{tab:app_dog_nine_seeds} reports the mean ± standard deviation for the ten post-hoc OOD detection methods under both testing protocols.

Across all methods, breed-aggregated training consistently yields near-random performance on OOD-breed detection, whereas breed-separated training achieves substantially higher AUROC. Similar improvements are observed under the OOD-object protocol. These results further support our original conclusion that training granularity plays a critical role in OOD detection performance.}

\begin{table*}[t]
\centering
\caption{\revise{AUROC (\%) of OOD detection methods on ImageNet Dogs under nine repeated experiments.}}
\label{tab:app_dog_nine_seeds}
\resizebox{0.7\textwidth}{!}{
\renewcommand{\arraystretch}{1.2}
\begin{tabular}{c|c|c|c|c}
\toprule
Testing Protocol                & \multicolumn{2}{c|}{OOD-object} & \multicolumn{2}{c}{OOD-breed} \\
\midrule
Training Granularity            & aggregated & separated & aggregated & separated \\
\midrule
MSP~\cite{MSP_2017}           & 80.9$\pm$0.5 & 95.2$\pm$0.4 & 51.6$\pm$0.5 & 73.9$\pm$0.5 \\
ODIN~\cite{ODIN_2018}         & 80.4$\pm$0.5 & 96.4$\pm$0.3 & 52.8$\pm$0.7 & 75.3$\pm$0.4 \\
EBO~\cite{EBO_2020}           & 80.8$\pm$0.5 & 97.7$\pm$0.2 & 52.5$\pm$0.7 & 76.2$\pm$0.5 \\
GradNorm~\cite{GradNorm_2021} & 60.6$\pm$1.1 & 94.3$\pm$0.7 & 50.3$\pm$0.5 & 59.9$\pm$1.7 \\
RMDS~\cite{RMDS_2021}         & 80.4$\pm$0.4 & 93.3$\pm$0.5 & 50.8$\pm$0.4 & 76.3$\pm$0.5 \\
KNN~\cite{KNN_2022}           & 74.7$\pm$0.9 & 97.9$\pm$0.3 & 50.4$\pm$0.6 & 75.2$\pm$0.7 \\
DICE~\cite{DICE_2022}         & 76.4$\pm$0.7 & 97.0$\pm$0.4 & 51.1$\pm$0.7 & 71.1$\pm$0.8 \\
ASH~\cite{ASH_2023}           & 76.3$\pm$1.7 & 97.3$\pm$0.3 & 51.8$\pm$1.0 & 70.1$\pm$1.7 \\
Relation~\cite{Relation_2023} & 79.2$\pm$0.7 & 96.9$\pm$0.2 & 51.7$\pm$0.9 & 73.6$\pm$0.8 \\
SCALE~\cite{SCALE_2024}       & 77.9$\pm$0.7 & 97.8$\pm$0.3 & 51.8$\pm$0.8 & 75.0$\pm$1.2 \\
\bottomrule
\end{tabular}
}
\end{table*}

\end{document}